\documentclass[10pt,twocolumn,letterpaper]{article}

\usepackage[pagenumbers]{cvpr} 
\usepackage{tabularx, multirow}
\usepackage{float}
\usepackage[ruled,vlined]{algorithm2e}

\definecolor{cvprblue}{rgb}{0.21,0.49,0.74}
\usepackage[pagebackref,breaklinks,colorlinks,allcolors=cvprblue]{hyperref}

\def\confName{CVPR}
\def\confYear{2027}

\title{Gaspard: Expressive unified speech and gesture synthesis with continuous autoregressive flow-matching and decoupled classifier-free guidance}

\author{
Téo Guichoux$^{1,2}$ \quad
Théodor Lemerle$^{2}$ \quad
Catherine Pelachaud$^{1,3}$ \quad
Nicolas Obin$^{2}$\\
$^{1}$ISIR \quad
$^{2}$IRCAM \quad
$^{3}$CNRS\\
Sorbonne University, Paris, France\\
{\small Corresponding author: \tt\small teo.guichoux@sorbonne-universite.fr}
}

\begin{document}
\maketitle
\begin{abstract}
Human communication involves highly coordinated multimodal signals such as speech and co-speech gestures. Existing approaches for generating both modalities commonly follow a cascaded pipeline, where speech is generated first and subsequently used to drive gesture generation; speech generation is therefore independent of gestures. We introduce Gaspard, a unified speech--gesture generation model based on continuous autoregressive generation with next-step flow-matching predictions, and trained on Seamless Interaction, a large-scale multimodal dataset. The heterogeneous temporal properties of speech and gesture motivate modality-specific prediction heads and decoupled classifier-free guidance over the two generation streams. This guidance rebalances the influence of speech and gesture at inference time, improving expressivity while mitigating dynamics collapse in continuous autoregressive gesture generation. Objective and perceptual evaluations show that Gaspard is a strong gesture generation model against unimodal baselines and that it produces highly synchronous speech and gestures compared to a matched cascaded system. 
\end{abstract}
    
\section{Introduction}
\label{sec:intro}
In human communication, speech and co-speech gestures are highly coordinated signals exhibiting bidirectional dependencies: speech influences gestures and gestures influence timing \cite{mcneill1992hand, kendon1980gesture}. For instance, gesture dynamics may be associated with changes in speech prosody, such as rhythm, pauses, pitch, or emphasis \citep{kendon1997gesture, Wagner2014GestureSpeech}. Despite such multimodal coupling, existing approaches for generating both modalities commonly rely on separate generative pipelines, where speech is first generated from text and subsequently used to condition gesture generation \cite{nagy2021framework}. This design has several limitations. First, its cascaded nature requires two successive inference processes, introducing additional computational cost for real-time human-machine interactions. Second, the two modalities are modeled with a directed dependency: text drives speech, and speech drives gestures. As a consequence, gestures can adapt to speech, but cannot condition speech generation, contradicting psycholinguistic evidence of speech and gesture co-production. Unified generation therefore provides an attractive alternative, where speech and gestures are generated within the same model and can both contribute to a shared multimodal communicative state.


Unified speech–gesture generation, however, is not simply a matter of combining two generative processes. Speech and gestures exhibit fundamentally different temporal dynamics: speech contains rapidly varying acoustic information, whereas gestures evolve more slowly, exhibit strong temporal correlations, and often contain extended periods of limited motion. When modeled jointly, these heterogeneous dynamics can create asymmetric autoregressive dependencies, where excessive reliance on previously generated gestures progressively suppresses motion dynamics. Preserving expressive gestures while maintaining cross-modal coordination is therefore a central challenge for unified generation.

In this work, we introduce \textbf{Gaspard} (Gesture and Speech autoregressive diffusion), a continuous autoregressive model for the unified synthesis of speech and co-speech gestures from text. Gaspard performs next-step prediction in continuous speech and gesture latent spaces, with each next-step distribution modeled using flow matching.

Gaspard addresses the challenges of unified speech–gesture synthesis by jointly modeling two modalities with fundamentally different temporal dynamics. We show that these asymmetries give rise to imbalanced autoregressive feedback, leading to progressive motion collapse where gesture autoregressive history overpowers cross-modal conditioning. Inspired by compositional guidance \cite{liu2022compositional, rouard2024audio}, we introduce a \emph{decoupled classifier-free guidance} strategy applied to jointly generated speech and gesture streams. Unlike prior formulations that combine multiple conditioning signals for a single generated stream, our method separately regulates self-modal autoregressive feedback and cross-modal conditioning in both modalities, preserving expressive gesture dynamics while enabling tightly coordinated speech--gesture generation.



Through objective and perceptual evaluations, we show that Gaspard enables controllable and expressive speech and gesture synthesis while generating both modalities in real time. We further compare Gaspard with recent unimodal gesture-generation baselines \cite{nagy2026genea} and with a matched cascaded system, investigate the role of modality-specific guidance, and show that generated speech is sensitive to changes in gesture dynamics.

Our main contributions can be summarized as follows:
\begin{itemize}[itemsep=0.15em, topsep=0.2em, leftmargin=*, label=\textbullet]
    \item We introduce a continuous autoregressive next-step flow-matching model for joint speech--gesture generation. 
    \item We identify and characterize an autoregressive feedback imbalance in unified generation, where excessive reliance on gesture history leads to progressive motion dynamics collapse.
    \item We adapt compositional classifier-free guidance to the autoregressive dependencies of jointly generated speech and gestures. Unlike conventional formulations that independently weight multiple conditions for a single generated stream, our \emph{decoupled CFG} separately regulates self-modal and cross-modal feedback in both generation streams, mitigating gesture dynamics collapse while strengthening speech--gesture coordination.
\end{itemize}

Example videos are provided at our project page: \href{https://tguichoux.github.io/papers/Gaspard}{%
    \textsf{https://tguichoux.github.io/papers/Gaspard}%
}

\section{Related Work}
\label{sec:sota}
\paragraph{Continuous autoregressive generation.}
Autoregressive generation of continuous data, such as speech or motion, is often performed through data discretization \cite{chen2025valle,borsos2023audiolm, guichoux2026gelina, liu2024emage, yi2023generating}. In this paradigm, VQ-VAEs \cite{oord2017neural} are used to quantize continuous signals into discrete codes, and the resulting code sequences are treated as tokens in an autoregressive language model. However, it also introduces an information bottleneck caused by compression and quantization, which can limit reconstruction quality \cite{sun2026multimodal}. 

Motivated by the success of continuous diffusion-based models, several recent works \cite{li2024autoregressive, sun2026multimodal,turetzky2024speech,rouard2026continuous, peng2026vibevoice, xiao2025motionstreamer} have shifted toward continuous latent representations for autoregressive generation. In these approaches, autoregressive generation is typically performed through next-step diffusion over continuous codes. In particular, an autoregressive backbone predicts a conditioning representation from the preceding sequence, while a diffusion head generates the next continuous latent code.

\paragraph{Dynamics collapse.}
A common issue in autoregressive human motion synthesis is dynamics collapse, where generated motion becomes repetitive, converges toward average poses, or drifts away from the latent manifold, a phenomenon often associated with regression to the mean \cite{Nyatsanga_2023}. One approach to mitigating this issue is to reduce the model's reliance on autoregressive motion history. During training, \cite{henter2020moglow} employ high motion dropout, while \cite{kucherenko2020gesticulator} combine non-autoregressive pre-training with scheduled teacher forcing for co-speech gesture generation. At inference time, \cite{rui2024taming} explore reducing motion-history guidance to improve style controllability in character animation. Gaspard builds on this principle by using classifier-free guidance at inference time to balance autoregressive gesture history and speech conditioning in joint speech--gesture generation.

\paragraph{Unified speech and gesture synthesis.}
A growing body of work investigates joint modeling of speech and co-speech gestures. Tacotron-ISG \cite{wang2021integrated} first showed that intermediate TTS representations can effectively condition gesture generation, while Diff-TTSG \cite{mehta2023diff} uses shared text representations to drive speech and gesture diffusion models. Closer coupling has also been explored by directly modeling both modalities within the same generative process. GlowTTS-ISG \cite{wang2021integrated} jointly generates acoustic and motion features through a flow-based decoder, while Match-TTSG \cite{mehta2024unified} adopts a single flow-matching model for speech and gesture synthesis. Recently, several works have explored full-duplex speech and gesture generation \cite{mughal2026miburi, wang2026fullduplex, nagano2026dyaplex}. Unlike the present work, these approaches focus on modeling dyadic interactions. However, ELLSA~2 \cite{wang2026fullduplex} also employs classifier-free guidance to enhance gesture expressivity.
Closer to our work, Gelina \cite{guichoux2026gelina} is an autoregressive model that jointly generates discrete speech and gesture tokens through a single interleaved sequence. Its discrete gesture representation also required a flow-matching decoder to recover smooth motion, while the 75~Hz speech-token rate led to long autoregressive sequences that hindered real-time inference.
These works establish unified speech--gesture generation as a viable approach, but largely leave unexplored the cross-modal dependencies that emerge during joint generation, including how speech and gesture histories differently influence subsequent predictions.

\section{Method}
\label{sec:method}

\begin{figure*}[!htb]
    \centering
    \includegraphics[width=0.78\textwidth]{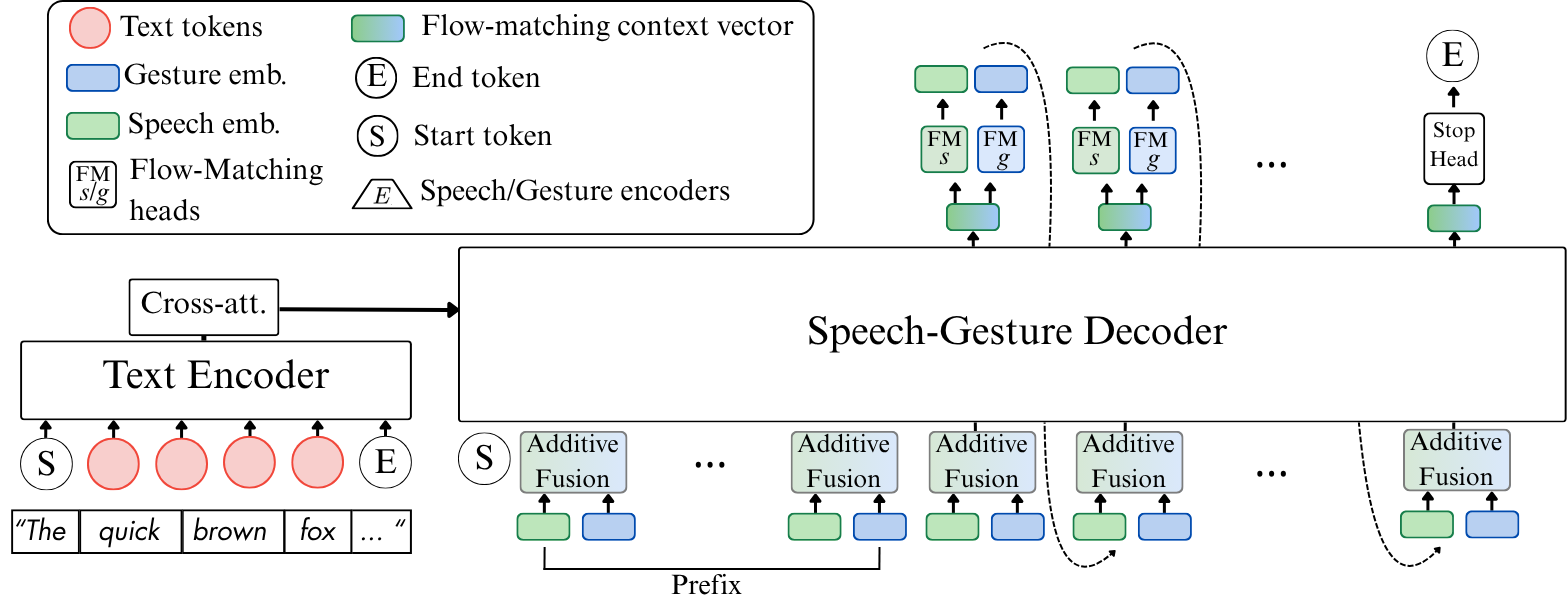}
    \caption{Gaspard generates speech and gesture continuous embeddings from text using modality-specific flow-matching heads. Speech and gesture embeddings are fused additively after modality-specific projection. At inference, a short reference prefix is provided to condition the generation, allowing both voice and gesturing style cloning.}
    \label{fig:gaspard}
\end{figure*}

Gaspard generates speech and co-speech gestures from a text input, using an encoder-decoder transformer architecture. Generation can further be conditioned on an example prefix. Autoregressive prediction is performed with two separate flow-matching heads that synthesize the next-step speech and gesture embeddings. An overview of the Gaspard architecture is provided in Figure~\ref{fig:gaspard}.

\subsection{Aligning speech and gesture representations}
\label{sec:aligninn_rep}

Speech waveform and gesture sequences are encoded with separate VAEs. Specifically, we use Z-Codec \citep{zcodec} for speech encoding; given a speech input $X^s \in \mathbb{R}^{M \times 1}$, Z-Codec encodes it into a sequence of latent embeddings denoted $Z^s \in \mathbb{R}^{d \times N}$.
For gesture representations, we use a fully convolutional $\sigma$-VAE that takes as input a gesture sequence $X^g \in \mathbb{R}^{r \times J \times L}$ and encodes it into a latent embedding sequence $Z^g \in \mathbb{R}^{d' \times N}$. Both speech and gesture VAEs operate at the same latent frame rate, which simplifies downstream autoregressive modeling.

\begin{figure*}[t]
    \centering
    \includegraphics[width=0.84\textwidth]{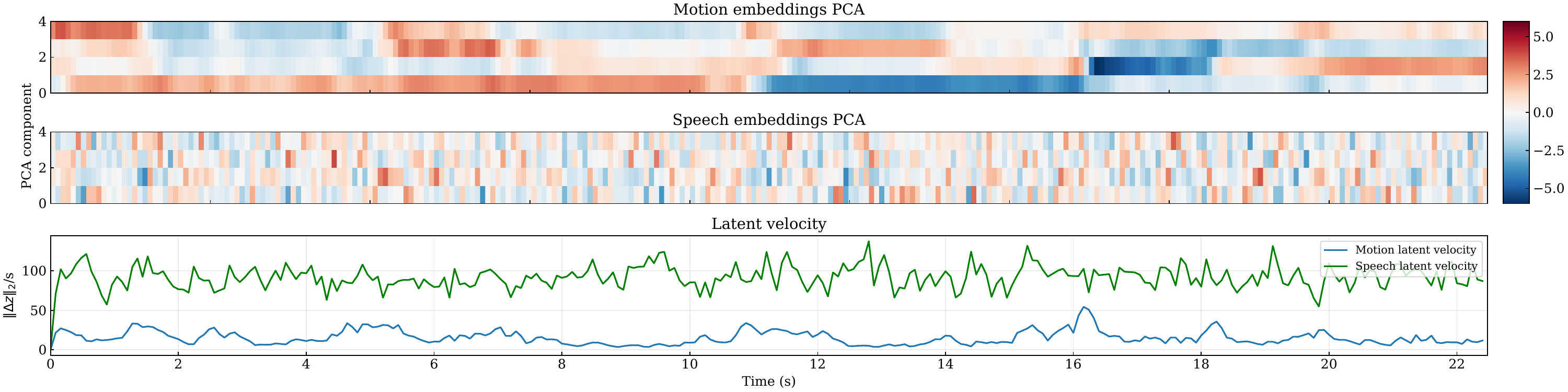}
    \caption{\textit{Top and middle}: PCA views of gesture and speech latent sequences at 12.5 Hz. Gesture and speech embedding sequences exhibit different properties. Gesture latent sequences (top) are smooth and slow-evolving, as demonstrated by repetitive patterns. In contrast, the speech latent sequence is fast-varying and discontinuous. \textit{Bottom:} Latent velocity of speech and gesture sequences. Velocity is computed as the difference between consecutive steps: $v^{s|g} = |z^{s|g}_{t+1} - z^{s|g}_{t}|$. Speech latent velocity (green) is much higher and more stable than the gesture latent velocity (blue), illustrating the different nature of both signals.}
    \label{fig:latent_signal}
    \label{fig:varying_signal}
\end{figure*}

Speech and gesture signals, although represented in a learned latent space, exhibit different properties. Speech embeddings, for example, form a fast-varying signal, whereas gesture embeddings form a slower and more conservative signal with many repetitive patterns, for example when a limb is not used for gesturing. Figure~\ref{fig:varying_signal} illustrates the different properties of speech and gestures.

\subsection{Continuous autoregressive speech-gesture generation}

Gaspard is trained to autoregressively generate a sequence of speech and gesture embeddings conditioned on textual input and next-step prediction is performed through flow-matching sampling. Voice and gestural style cloning can be performed via prefix continuation.


At each autoregressive step, the previously generated speech and gesture embeddings are first projected to the model dimension and fused additively on the input side. The resulting fused sequence $e_{<t}$ is processed by the autoregressive backbone $B_\theta$, together with the encoded text representation, to produce a shared context vector:
\begin{equation}
c_t = B_\theta(e_{<t}, x).
\end{equation}
This context vector conditions two modality-specific flow-matching heads, $h^s_\theta$ and $h^g_\theta$, parameterized as small MLPs following \cite{sun2026multimodal, peng2026vibevoice}, which model the next speech and gesture embeddings in continuous space.

For next-step prediction, the model samples the next embeddings by integrating two conditional velocity fields, one for speech and one for gesture. For each modality $m \in \{s,g\}$, inference starts from a Gaussian sample $z^m_{0,t} \sim \mathcal{N}(0, I)$ and integrates the ODE:
\begin{equation}
\frac{d z^m_{\tau,t}}{d\tau}
=
h^m_\theta(z^m_{\tau,t}, \tau, c_t),
\qquad
\tau \in [0,1],
\label{eq:gaspard_ode}
\end{equation}
where $\tau$ denotes the flow-matching sampling time. The generated next embedding is obtained at the end of the integration: $\hat{z}^m_t = z^m_{1,t}$.

The overall backbone architecture follows Gelina \cite{guichoux2026gelina}. Text is encoded by a non-causal GLA-based encoder \citep{yang2024gated} and conditions the autoregressive decoder through cross-attention, while the causal GLA-based decoder models the fused speech-gesture history. At each autoregressive step, a binary classification head predicts whether the generation should stop.

\paragraph{Training:}

Training is performed in a teacher-forcing next-step prediction regime, with a flow-matching objective. At each autoregressive step $t$, the backbone receives the ground-truth prefix $(z^s_{<t}, z^g_{<t})$ and outputs the context vector $c_t$. For each modality $m \in \{s,g\}$ the modality-specific flow-matching heads are trained to predict the target velocity field $ v^m_t= z^m_{1,t} - z^m_{0,t}$ from the intermediate state $z^m_{\tau,t}$, the flow-matching time $\tau$, and the autoregressive context $c_t$:
\begin{equation}
\mathcal{L}^{m}_{\mathrm{FM}}
=
\mathbb{E}_{t,\tau,z^m_{0,t}}
\left[
||
v^m_t
-
h^m_\theta(z^m_{\tau,t}, \tau, c_t)
||_2^2
\right],
\qquad
m \in \{s,g\}.
\end{equation}
The final Gaspard objective is a weighted combination of the speech and gesture flow-matching losses:
\begin{equation}
\label{eq:gaspard_objective}
\mathcal{L}
=
\lambda_s \mathcal{L}^{s}_{\mathrm{FM}}
+
\lambda_g \mathcal{L}^{g}_{\mathrm{FM}}
+ 
\lambda_{\mathrm{stop}}\mathcal{L}_{\mathrm{stop}}
\end{equation}
where $\lambda_s$ and $\lambda_g$ are modality-specific loss weights, and $\mathcal{L}_{\mathrm{stop}}$ is the binary Cross-Entropy loss term controlling the stop-prediction head.


\subsection{Decoupled classifier-free guidance for expressive multimodal synthesis}
\label{sec:decoupled_cfg}
Classifier-free guidance is a common strategy to encourage fidelity to a control signal in diffusion-based generative models, and has been widely used in continuous autoregressive next-step diffusion settings \citep{ke2026hyperspherical, sun2026multimodal, rouard2026continuous, lemerle2026unbounded}. 
By masking one input at a time, we can obtain context vectors that remove the contribution of specific modalities. We denote by $c_t^{\neg G}$ and $c_t^{\neg S}$ the contexts obtained by masking gesture and speech, respectively:
\begin{equation}
\begin{aligned}
c_t^{\neg G}
&=
B_\theta
\left(
\mathrm{Fuse}(z^s_{<t}, \varnothing), x
\right),
\\
c_t^{\neg S}
&=
B_\theta
\left(
\mathrm{Fuse}(\varnothing, z^g_{<t}), x
\right).
\end{aligned}
\end{equation}

Given a flow-matching head $h_\theta$,and $ m \in \{S, G\}$, the corresponding velocity fields are:
\begin{equation}
v
=
h_\theta(c_t, z_\tau, \tau),
\qquad
v^{\neg m}
=
h_\theta(z_\tau, \tau, c_t^{\neg m})
\end{equation}

Inspired by \cite{rouard2024audio} and \cite{lemerle2026unbounded}, we define a chained multimodal CFG by progressively rebalancing the contribution of each modality. For each flow head $h_\theta^m$, and starting from the full conditional velocity $v$, the guided velocity is updated as:
\begin{equation}
\begin{aligned}
v_{\mathrm{cfg}}
&\leftarrow
v,
\\
v_{\mathrm{cfg}}
&\leftarrow
v^{\neg G}
+
\gamma_g^m
\left(
v_{\mathrm{cfg}} - v^{\neg G}
\right),
\\
v_{\mathrm{cfg}}
&\leftarrow
v^{\neg S}
+
\gamma_s^{m}
\left(
v_{\mathrm{cfg}} - v^{\neg S}
\right),
\\
\end{aligned}
\end{equation}

\begin{figure}[H]
    \includegraphics[width=0.38\textwidth]{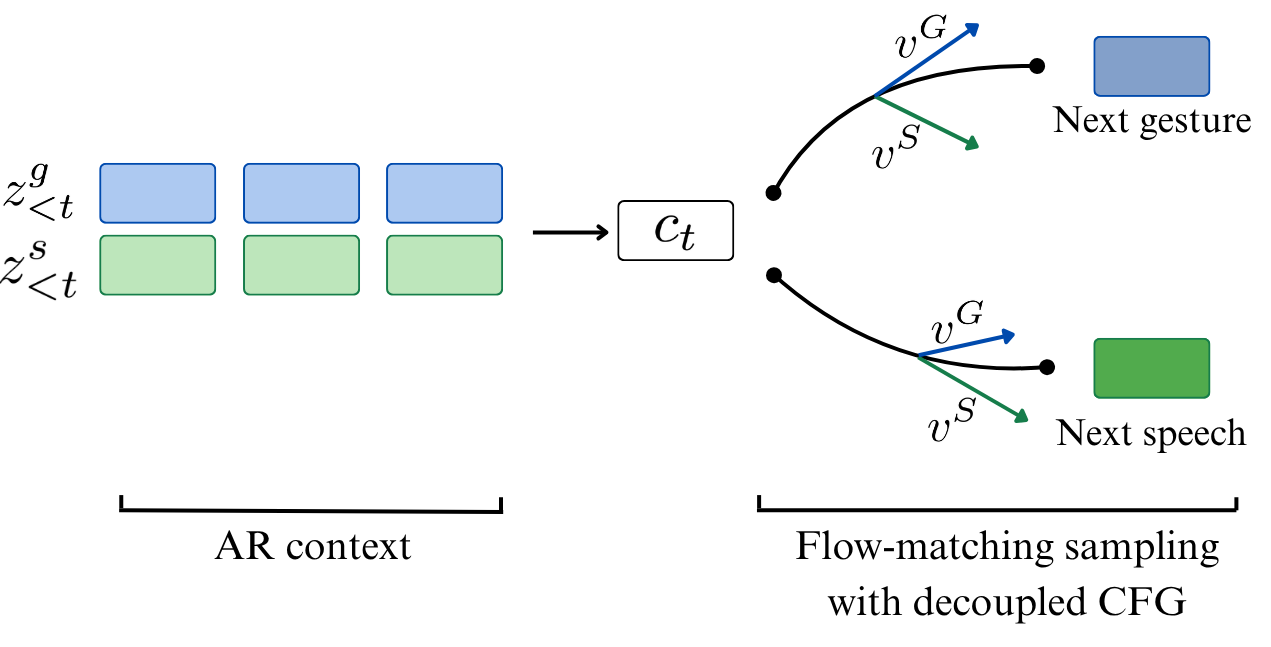}
    \caption{Overview of our decoupled CFG strategy. Next-step speech and gesture latents are sampled conditioned on the context vector $c_t$ predicted from the gesture and speech fused history, $z^g_{<t}$ and $z^s_{<t}$}
    \label{fig:cfg}
\end{figure}

Here, $\gamma_G$ and $\gamma_S$ control the contribution of gesture and speech, respectively. 
With $\gamma^G_g <1$, the difference between the full and gesture-masked predictions is reduced, reducing the influence associated with gesture history. In contrast, $\gamma^G_s >1$ increases the influence of the speech history on gesture prediction. Similarly, $\gamma^S_g$ controls the influence of gesture history on speech generation, while $\gamma^S_s$ controls the influence of the speech history.
The guidance scales assign distinct coefficients to the contribution of each modality, providing modality-specific control over self-modal and cross-modal conditioning. The corresponding guidance updates are composed sequentially, and in all experiments, gesture guidance is applied first, followed by speech guidance.\footnote{For fixed guidance scales, the chained formulation is algebraically equivalent to an explicit compositional CFG formulation with independently weighted gesture- and speech-conditioning terms. See Sup.~12.3.1 for discussion.} We found this chained formulation practical for hyperparameter tuning. We summarize the intuition behind each coefficient in Table~\ref{tab:cfg_coefficients}, and Figure~\ref{fig:cfg} illustrates the decoupled CFG strategy.

An overview of the autoregressive inference loop with our decoupled CFG mechanism is depicted in Algorithm~\ref{alg:gaspard_inference}.
\section{Experimental Setup}
\label{sec:exp_setup}
\label{sec:experimental_setting}
\paragraph{Data and training.}
We trained Z-Codec's PatchVAE, the gesture $\sigma$-VAE, and Gaspard on the Seamless Interaction dataset \cite{seamless_interaction}. The public dataset contains $4,008$~h of paired speech--gesture data from $4,284$ participants, together with transcriptions, and is divided into \textit{naturalistic} ($2,745$~h) and \textit{improvised} ($1,319$~h) interactions. Speech is resampled from $48$ to $24$~kHz to match Z-Codec. We retain upper-body SMPL-H motion, remove leg motion and root translation, convert relative joint rotations to Rot6D~\cite{zhou2019onthecontinuity}, and resample motion from $30$ to $25$~fps. Recordings are resegmented using the provided voice activity annotations, and we retain active-speaking segments only. Detailed split statistics, preprocessing parameters, and segmentation procedures are provided in Appendix~\ref{app:dataset_preprocessing}.

The backbone is a 246M-parameter encoder--decoder Transformer with gated linear attention, comprising a 6-layer text encoder, a 12-layer causal decoder, and a 1024-dimensional hidden space. Gaspard was trained for 200k steps on 2 H100 GPUs using bf16 mixed-precision, corresponding to approximately 40 h of training. The pretrained speech and gesture autoencoders are kept frozen during this training stage. At inference time, we use $15$ flow-matching sampling steps for both the speech and gesture heads. Full architecture and optimizer configurations are provided in Appendix~\ref{app:training_hyperparameters}.

\paragraph{Evaluation sets and baselines.}
\label{sec:clean_test_set}
We conduct our evaluation on \textit{two} evaluation subsets of the test split of the Seamless Interaction dataset. The comparison against gesture-generation baselines is performed on a manually curated subset used in the Genea Challenge \cite{nagy2026genea}, referred to as \textit{Genea-95} (95 segments). We also include an automatically curated subset, \textit{Seamless-416}, containing $416$ segments from $93$ speakers, for a total duration of approximately $2$~h~$35$~min. Evaluation recordings range from $15$ to $35$~s. For each speaker, a unique separate segment of at most $10$~s is selected as a speech and gesture prefix for generation.
We provide additional details regarding segment selection, curation, and other processing steps in Appendix~\ref{app:clean_subset}.

At the time of the research and to the best of our knowledge, there exist no open-source unified speech and gesture generation models trained on the Seamless Interaction dataset. Because of differing pose representations (rotations/positions, considered limbs, etc.), adapting existing models to the dataset would require substantial engineering efforts. We therefore compare Gaspard against recent unimodal speech-driven gesture generation models: \textit{UNICAMP} \cite{sanchez2026unified}, \textit{GestFlow} \cite{guillen2026gestflow}, \textit{DyaSync-Sem} \cite{fang2026dyasync}, and \textit{UM-FERI} \cite{crnek2026umferi}, trained on Seamless Interaction and evaluated in the Genea Challenge 2026 \cite{nagy2026genea}. These baselines use ground-truth speech originating from the dataset, whereas Gaspard generates both the speech and the gestures from text. We additionally compare Gaspard against the teacher-version of PocketTTS \cite{rouard2026continuous}, a 350M-parameter continuous autoregressive TTS model, which we trained on the Seamless Interaction dataset. We finally construct a \textit{Cascaded} baseline combining a unimodal text-to-speech model with a speech-driven gesture generation model, which follows the same training recipe as Gaspard. Its TTS component corresponds exactly to Gaspard, without gesture embedding inputs and gesture flow-matching head. The speech-driven gesture generation component follows the Gaspard architecture, except that the speech objective weight, $\lambda_{\mathrm{s}}$ in Equation~\ref{eq:gaspard_objective}, is set to $0.0001$ and kept only for regularization purposes. All remaining architectural components and training settings are kept unchanged. 
We provide additional details about the baseline models in Appendix~\ref{app:baselines}.

\paragraph{Common evaluation protocol.}
We evaluate speech and gesture synthesis separately. Gesture metrics include Fr\'echet Gesture Distance (FGD) \cite{yoon2020speech}, Beat Consistency (BC) \cite{li2021learn}, and $\ell_1$-Diversity \cite{liu2024emage}. To complement FGD and inspired by \cite{ng2024audio2photoreal}, we introduce kinetic Fréchet Gesture Distance, denoted by $\mathrm{FGD}_k$. Rather than comparing latent features directly, $\mathrm{FGD}_k$ compares distributions of first-order temporal variations in the latent space. Appendix~\ref{app:metric_details} describes and validates the metric.
FGD measures pose-distribution similarity, whereas $\mathrm{FGD}_k$ aims to capture motion dynamics similarities. BC measures speech--gesture synchrony, and diversity measures motion variability. For speech, we report Word Error Rate (WER), UTMOS \cite{saeki2022utmos}, and Speaker Similarity (SS) computed with WavLM \cite{chen2022wavlm}. We also report real-time factor (RTF).

We conducted \textit{two} perceptual studies, one for comparison against gesture generation baselines and one for our ablation study. For both studies, all stimuli were rendered in Blender using the 3D scene and avatar representations used in the Genea Leaderboard \citep{kucherenko2023genea}. The average duration of the stimuli is 20~s. Participants recruited through Prolific evaluated $30$ randomly sampled stimuli on a $1$--$5$ Likert scale and completed two attention checks. Human-likeness of gestures was evaluated from muted videos; speech--gesture synchronization was evaluated from audiovisual stimuli. The ablation study additionally evaluated voice human-likeness from audio alone. The evaluation subsets and participant groups differed between studies, as specified below and further detailed in Appendix~\ref{app:clean_subset}. For both studies, participant-level MOS were analyzed with repeated-measures ANOVA and Holm-corrected pairwise tests.
\section{Results}
\label{sec:results}
\subsection{Comparison to gesture-generation baselines}
\begin{table}[t]
    \centering
    \scriptsize
    \setlength{\tabcolsep}{2.9pt}
    \resizebox{\linewidth}{!}{%
    \begin{tabular}{lcccc}
        \toprule
        Method & FGD $\downarrow$ & $\mathrm{FGD}_k$ $\downarrow$ & BC $\sim$ & Div. $\sim$ \\
        \midrule
        Human          & -- & -- & $0.63$ & $26.15$ \\
        \midrule
        DyaSync-S & $31.00 \pm 0.12$ & $3.40 \pm 0.04$ & $0.82$ & $18.46 \pm 0.05$ \\
        GestFlow    & $\underline{8.84 \pm 0.35}$ & $1.49 \pm 0.08$ & $0.91$ & $19.31 \pm 0.17$ \\
        UM-FERI     & $11.01 \pm 0.14$ & $\mathbf{1.02 \pm 0.04}$ & $\underline{0.74}$ & $\mathbf{25.33 \pm 0.14}$ \\
        UNICAMP     & $11.93 \pm 0.78$ & $2.04 \pm 0.23$ & $0.86$ & $16.35 \pm 0.41$ \\
        Gaspard (ours)     & $\mathbf{6.56 \pm 0.16}$ & $\underline{1.35 \pm 0.21}$ & $\mathbf{0.67}$ & $\underline{21.65 \pm 0.54}$ \\
        \bottomrule
    \end{tabular}}
    \caption{Evaluation results. Best results are shown in bold and second-best results are underlined. For BC, performance is ranked by proximity to the ground-truth value. $\pm$ indicates $95\%$ confidence intervals.}
    \label{tab:genea_results}
\end{table}
We present the comparison with unimodal gesture-generation models on \textit{Genea-95} in Table~\ref{tab:genea_results}. Gaspard obtains the lowest FGD ($6.56$) and a BC score comparable to human recordings ($0.67$ vs. $0.63$), while ranking second in $\mathrm{FGD}_k$ and diversity, behind \textit{UM-FERI} for both aspects.

Table~\ref{tab:pockettts} contains the results of the speech metrics of PocketTTS and Gaspard.
\begin{table}[t]
    \centering
    \scriptsize
    \setlength{\tabcolsep}{4.0pt}
    \resizebox{\linewidth}{!}{%
    \begin{tabular}{lcccc}
        \toprule
        Method & WER (\%) $\downarrow$ & UTMOS $\uparrow$ & SS $\uparrow$ \\
        \midrule
        Human          & -- & $2.88 \pm 0.05$ & --\\
        PocketTTS      & $\mathbf{4.69 \pm 1.15}$ & $\mathbf{3.65 \pm 0.05}$ & $0.70 \pm 0.015$ \\
        Gaspard (ours) & $6.38 \pm 1.35$ & $2.69 \pm 0.05$ & $\mathbf{0.77 \pm 0.015}$\\
        \bottomrule
    \end{tabular}}
    \caption{Speech generation results on the Genea-95 subset. WER denotes word error rate, UTMOS the predicted naturalness score, and SS the speaker similarity. Best results among generated samples are shown in bold. $\pm$ indicates $95\%$ confidence intervals.}
    \label{tab:pockettts}
\end{table}
PocketTTS achieves lower WER (4.69\% versus 6.38\%) and higher predicted naturalness (3.65 versus 2.69), whereas Gaspard achieves higher speaker similarity (0.77 versus 0.70).

\subsubsection{Perceptual study protocol}
For this comparison, we randomly sampled $30$ segments from \textit{Genea-95}, using the corresponding baseline outputs provided by the Genea Challenge organizers \cite{nagy2026genea}. We considered six conditions: \textit{Human}, \textit{Gaspard}, and the four gesture-generation baselines. We recruited $60$ participants, divided between gesture human-likeness and audiovisual synchronization evaluations. One participant was excluded after failing attention checks and was replaced.

\begin{figure}[]
    \centering
    \includegraphics[width=1.0\linewidth]{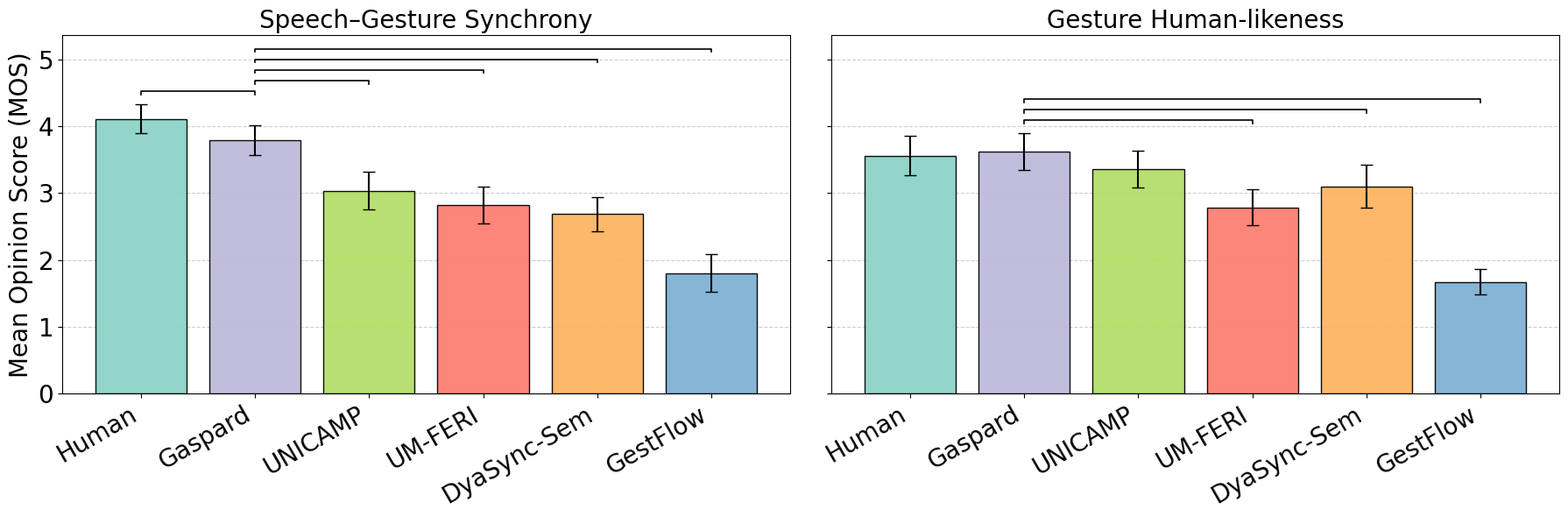}
    \caption{Results of the user study across the six conditions, reported as MOS with $95\%$ confidence intervals. Horizontal brackets indicate statistically significant differences ($p<0.05$). \textit{Left}: speech--gesture synchrony. \textit{Right}: gesture human-likeness. We only indicate significant differences w.r.t Gaspard for clarity.}
    \label{fig:user_study_results_genea}
\end{figure}
\subsubsection{Perceptual study results}
Gaspard achieved the highest speech--gesture synchrony ratings among the generative systems, significantly outperforming all competing models after Holm correction. Gaspard also obtains the highest mean motion human-likeness score, with no statistically significant difference from ground-truth motion, and its advantage over the strongest competing model, \textit{UNICAMP}, is not statistically significant.

\subsection{Ablation studies}
We conducted our ablation studies on the \textit{Seamless-416} evaluation subset.
\subsubsection{Decoupled classifier-free guidance}
\label{sec:impact_cfg}

\paragraph{Guidance effects on motion and speech.}

\begin{figure}[!htb]
    \centering
    \includegraphics[width=1.0\linewidth]{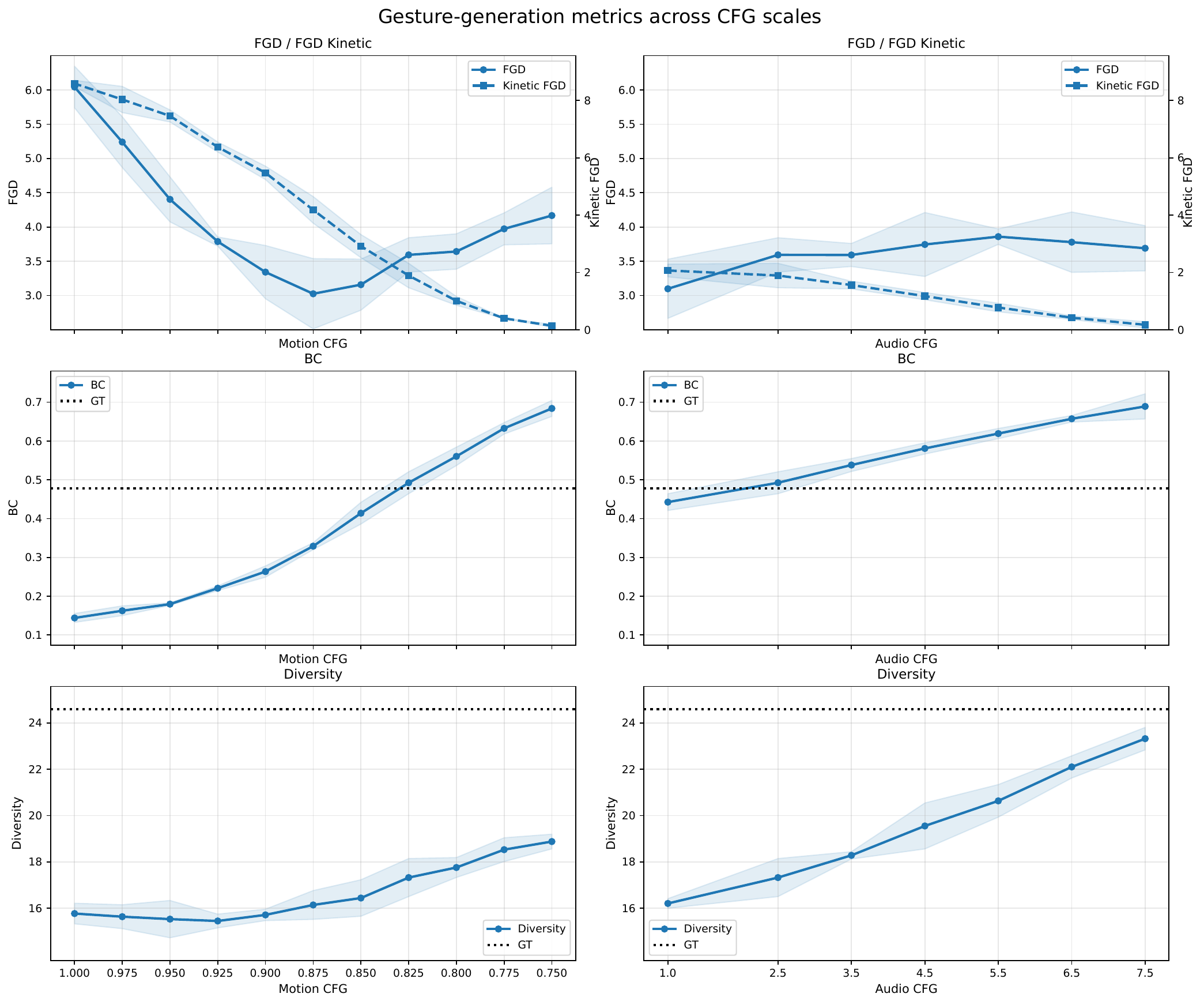}
    \caption{\textit{Left:} Gesture-history guidance $\gamma_g^G$ ($\gamma_s^G=2.5$).
\textit{Right:} Speech-history guidance $\gamma_s^G$ ($\gamma_g^G=0.825$).
Shaded regions show 95\% confidence intervals.}
    \label{fig:cfg_all}
\end{figure}

\begin{figure}[!htb]
    \centering

    \begin{subfigure}{1.0\linewidth}
        \centering
        \includegraphics[width=\linewidth]{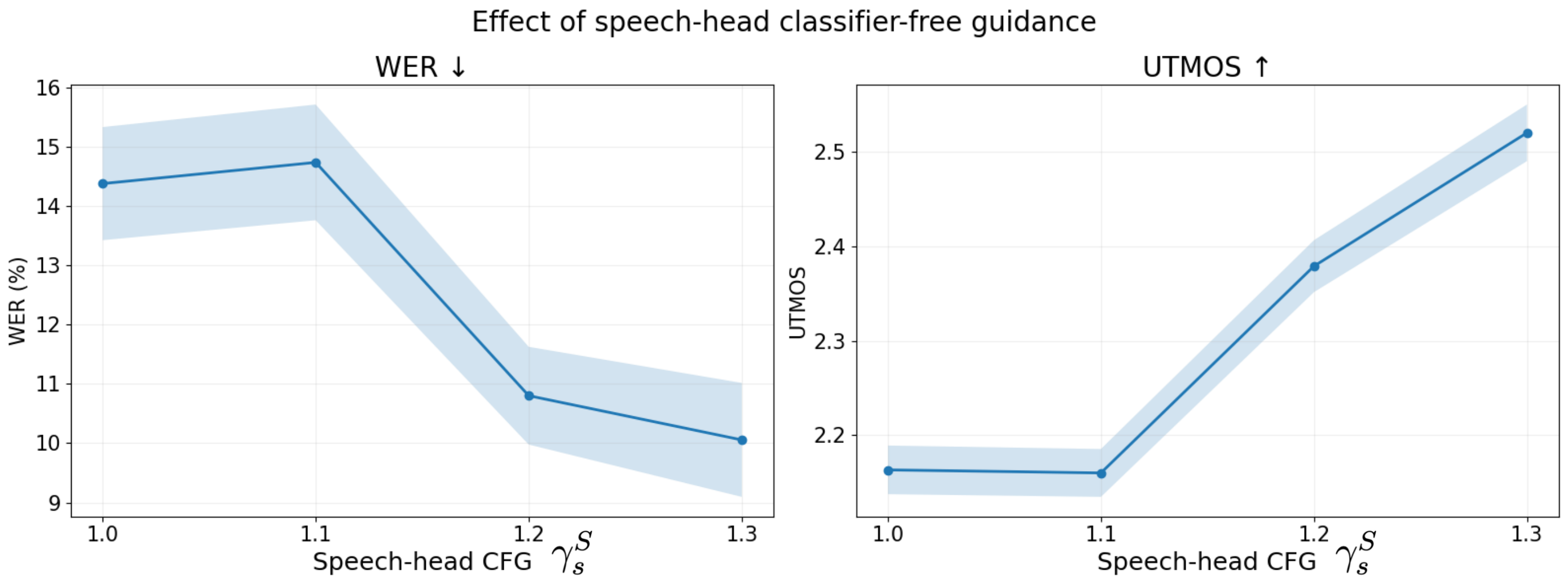}
        \caption{Varying $\gamma_s^S$. $\gamma^S_g$ is fixed to $1.0$}
        \label{fig:speech_head_cfg_gammas}
    \end{subfigure}

    \vspace{0.5em}

    \begin{subfigure}{1.0\linewidth}
        \centering
        \includegraphics[width=\linewidth]{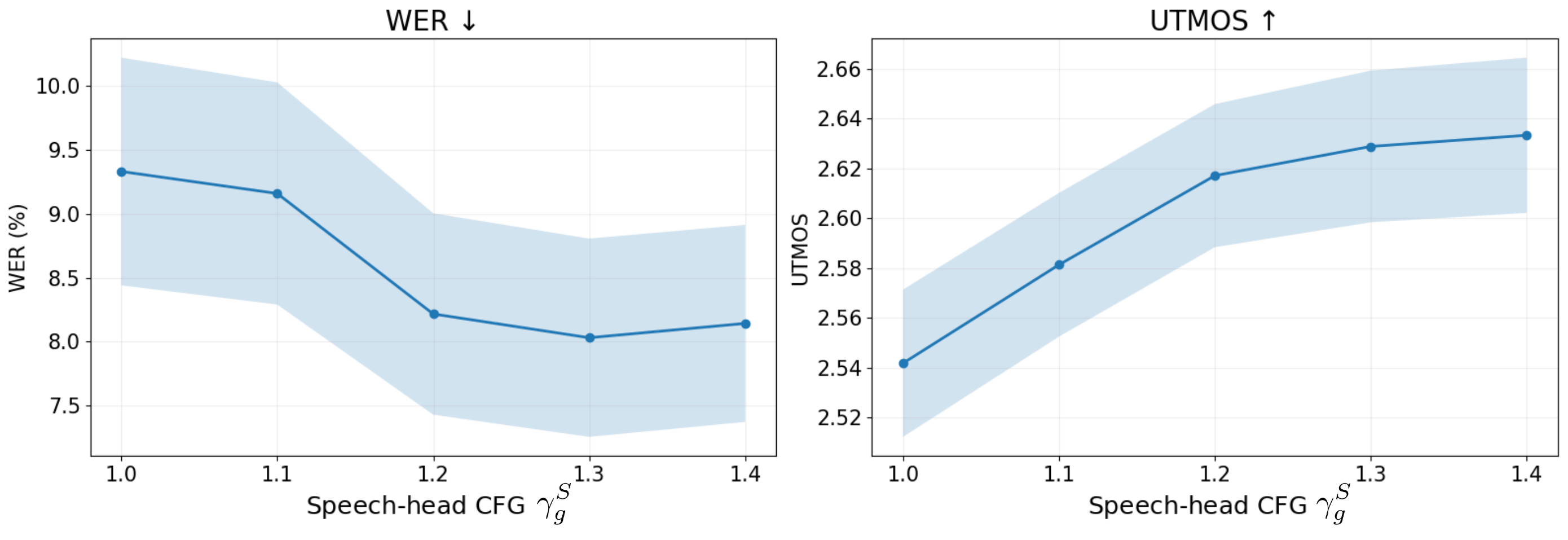}
        \caption{Varying $\gamma_g^S$. $\gamma^S_s$ is fixed to $1.3$}
        \label{fig:speech_head_cfg_gammag}
    \end{subfigure}

    \caption{Effect of classifier-free guidance on generated speech. \textit{Top:} speech-history guidance $\gamma_s^S$. \textit{Bottom:} gesture-history guidance $\gamma_g^S$.}
    \label{fig:speech_head_cfg}
\end{figure}

\begin{table*}[t]
\centering
\scriptsize
\caption{Comparison on \textit{Seamless-416} of no CFG, standard CFG ($\gamma^{S}=1.3$, $\gamma^{G}=0.75$), decoupled CFG ($\gamma_s^S=\gamma_g^S=1.3$, $\gamma_g^G=0.825$, $\gamma_s^G=6.5$), and \textit{Cascaded} ($\gamma^{S}=1.3$, $\gamma_g^G=0.825$, $\gamma_s^G=6.5$). $\pm$: 95\% CI.}
\label{tab:cfg_comparison}
\setlength{\tabcolsep}{3.9pt}
\renewcommand{\arraystretch}{1.1}
\begin{tabular}{lcccccccc}
\toprule
Configuration & FGD$\downarrow$ & $\mathrm{FGD}_k\downarrow$ & BC $\sim$ & Div. $\sim$ & WER$\downarrow$ & UTMOS$\uparrow$ & SS$\uparrow$ & RTF$\downarrow$ \\
\midrule
Human & -- & -- & 0.478 & 24.59 & -- & $2.21\pm0.16$ & -- & -- \\
No CFG & $8.77\pm2.06$ & $12.32\pm1.43$ & $0.046\pm0.023$ & $11.41\pm1.77$ & $29.51\pm1.44$ & $1.76\pm0.02$ & $0.73\pm0.09$ & 0.78 \\
Standard CFG & $4.34\pm1.18$ & $0.81\pm0.43$ & $0.761\pm0.05$ & $22.02\pm0.83$ & $21.57\pm1.02$ & $1.91\pm0.39$ & $0.74\pm0.01$ & 0.78 \\
Decoupled CFG & $3.71\pm1.71$ & $0.42\pm0.33$ & $0.657\pm0.043$ & $22.00\pm1.08$ & $8.03\pm0.75$ & $2.62\pm0.03$ & $0.75\pm0.01$ & 0.84 \\
Cascaded & $3.20\pm2.60$ & $1.16\pm0.05$ & $0.74\pm0.042$ & $22.85\pm0.73$ & $7.73\pm0.71$ & $2.68\pm0.03$ & $0.77\pm0.01$ & 1.04 \\
\bottomrule
\end{tabular}
\end{table*}
We compare decoupled CFG with standard CFG and no guidance on the clean Seamless evaluation set; the results are reported in Table~\ref{tab:cfg_comparison}. Decoupled CFG substantially reduces FGD, $\mathrm{FGD}_k$, and WER, and improves BC, diversity and UTMOS compared with standard CFG and no CFG, at the cost of slightly slower inference caused by the additional backbone forward passes.
Figure~\ref{fig:cfg_all} summarizes the variations in FGD, $\mathrm{FGD}_k$, BC, and $\ell_1$-Diversity for different combinations of $\gamma_g^G$ and $\gamma_s^G$. Decreasing the guidance with respect to the gesture history initially improves both FGD and $\mathrm{FGD}_k$, confirming that weaker gesture-history guidance mitigates static motion collapse. However, when $\gamma_g^G$ becomes too low, motion coherence deteriorates, as illustrated by the increase in FGD observed at $\gamma_g^G = 0.75$.
Figure~\ref{fig:speech_head_cfg} illustrates the effect of $\gamma_s^S$ on speech intelligibility and quality. Increasing $\gamma_s^S$ to $1.3$ reduces the WER ($14.37\% \rightarrow 10.05\%$), while increasing UTMOS ($2.16\rightarrow2.52$). Varying the gesture guidance $\gamma_g^S$ also improves average WER from $9.32\%$ at $\gamma^S_g=1.0$ to $8.03\%$ at $\gamma^S_g=1.3$, and UTMOS from $2.54$ to $2.63$ at $\gamma^S_g=1.4$.

\paragraph{Decoupled CFG for improved gesture dynamics.}

\begin{figure}[]
    \centering
    \includegraphics[width=1.0\linewidth]{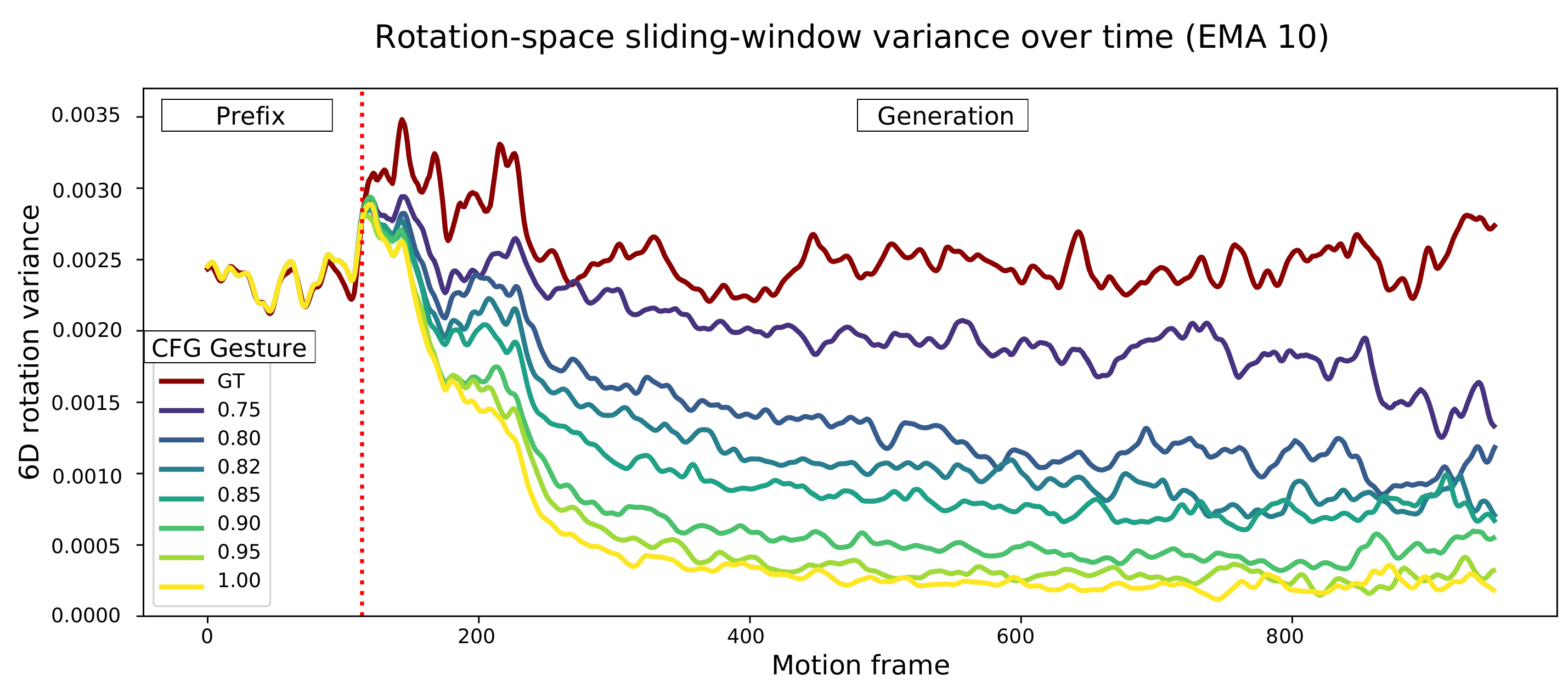}
    \caption{Impact of the gesture-history guidance scale $\gamma_g^G$ on the dynamics of the generated gestures.}
    \label{fig:guidance_collapse_motion}
\end{figure}

Figure~\ref{fig:guidance_collapse_motion} presents the impact of varying the gesture-history guidance scale $\gamma_g^G$, while keeping the speech guidance scale fixed at $\gamma_s^G = 2.5$, isolating the effects of variations in $\gamma_g^G$ while maintaining sufficiently strong speech conditioning. The portion preceding the vertical dotted line corresponds to teacher-forced predictions conditioned on the ground-truth prefix (1-step horizon), while the portion following the dotted line corresponds to free-running generation after the prefix. As shown in the figure, setting $\gamma_g^G = 1$ leads to a collapse of the generated dynamics, resulting in nearly static motion characterized by low variance. Progressively decreasing $\gamma_g^G$ mitigates this collapse and increases both the variability of the generated gestures, reaching velocity and variance levels comparable to those of the ground-truth motion, represented by the red curve. Teacher-forced generated gestures (the prefix at the left of dotted line) exhibit similar properties to the ground-truth gestures, suggesting that dynamics collapse is a self-reinforcing phenomenon during sequential generation.
Further decreasing the gesture-history guidance below $\gamma_g^G = 0.80$ degrades motion quality. In particular, values such as $\gamma_g^G = 0.75$ produce less human-like and increasingly jerky gestures, despite demonstrating variance similar to human recordings. This suggests that excessively weak conditioning on the gesture history prevents the next-step predictor from preserving sufficiently coherent motion trajectories.

\begin{figure}[]
    \centering
    \includegraphics[width=1.0\linewidth]{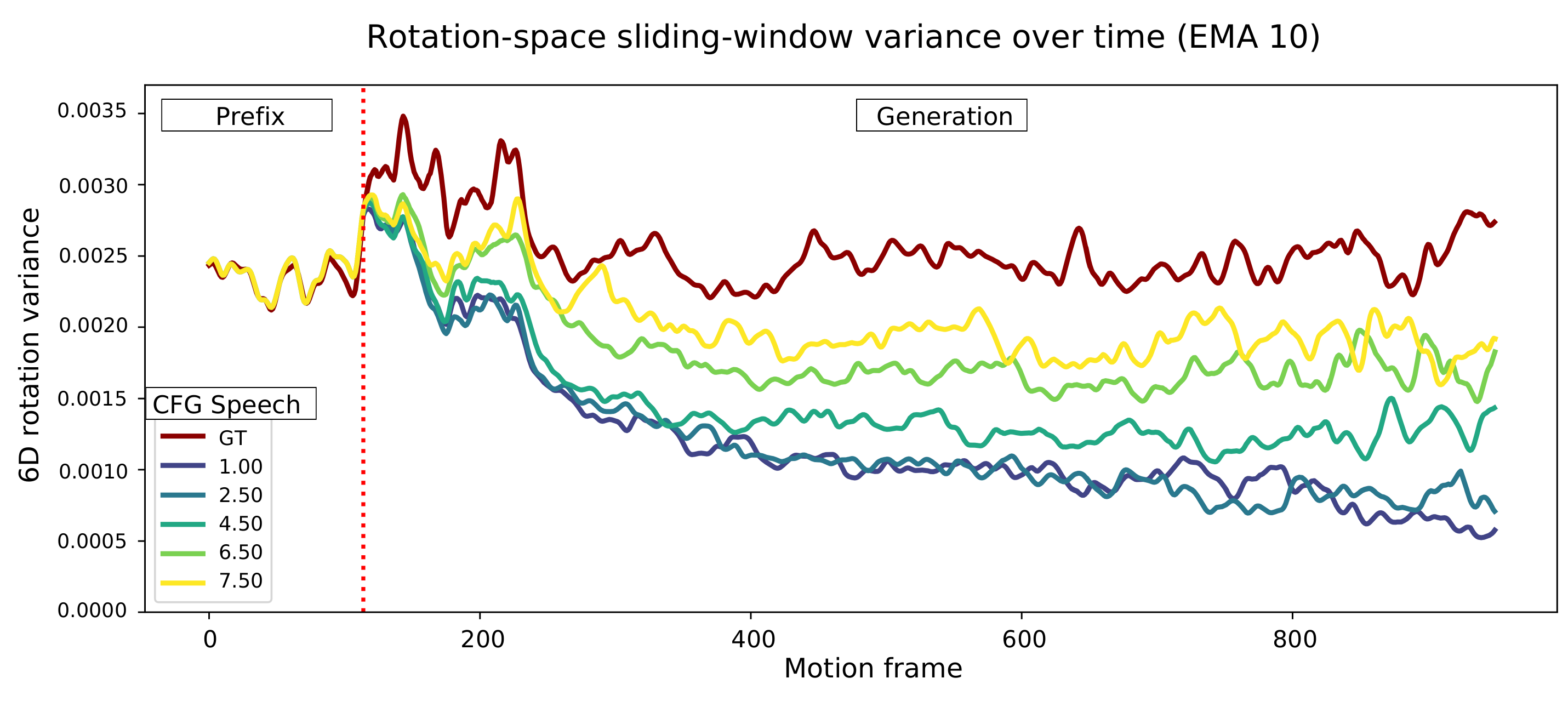}
    \caption{Impact of the speech guidance scale $\gamma_s^G$ on the dynamics of the generated gestures.}
    \label{fig:guidance_collapse_speech}
\end{figure}

Figure~\ref{fig:guidance_collapse_speech} presents the impact of varying the speech guidance scale $\gamma_s^G$, with the gesture-history guidance fixed at $\gamma_g^G = 0.825$. Increasing $\gamma_s^G$ increases the variability of the generated gestures. This behavior contrasts with that of $\gamma_g^G$: stronger speech guidance encourages the model to produce more pronounced motion variations in response to the speech signal.

\subsubsection{Unified versus cascaded generation}
\label{sec:cascaded_results}

We report the results of the \textit{Cascaded} ablation in Table~\ref{tab:cfg_comparison}. \textit{Unified} and \textit{Cascaded} exhibit different strengths across the gesture metrics. \textit{Cascaded} achieves a lower FGD and slightly higher diversity, whereas \textit{Unified} substantially improves $\mathrm{FGD}_k$ and produces a BC score closer to the ground-truth recordings. For speech, \textit{Cascaded} obtains a slightly lower WER and higher UTMOS and Speaker Similarity. Unified generation is more efficient, with an RTF of $0.84$ compared with $1.04$ for the cascaded approach.

To further investigate cross-modal coupling, we analyze the influence of gesture conditioning on generated speech rate. Varying the speed of the conditioning gestures produces correlated changes in speech rate and inter-pausal-unit structure, with a median correlation of $0.70$ and $-0.60$ respectively across multiple speakers and segments. This provides evidence that gesture context can influence speech realization in the unified model, which is impossible for speech-first cascaded architectures. Additional results and metric definitions are provided in Appendix~\ref{app:speech_adaptation}.

\subsubsection{Perceptual ablation}

\paragraph{User study protocol.}
We constructed an evaluation set of $24$ segments from $11$ speakers and recruited $120$ participants across three groups evaluating gesture human-likeness, voice human-likeness, and synchronization. The configurations were $\mathrm{G}_{s=6.5,g=0.825}$ ($\gamma^G_g=0.825$, $\gamma^G_s=6.5$), $\mathrm{G}_{s=1,g=0.825}$ ($\gamma^G_g=0.825$, $\gamma^G_s=1.0$), $\mathrm{G}_{s=2.5,g=1}$ ($\gamma^G_g=1.0$, $\gamma^G_s=2.5$), and \textit{Cascaded}, together with \textit{Human}. Seven participants were excluded after failed attention checks and replaced. 

\paragraph{Ablation study perceptual results.}
\begin{figure}[!htb]
    \centering
    \includegraphics[width=1.0\linewidth]{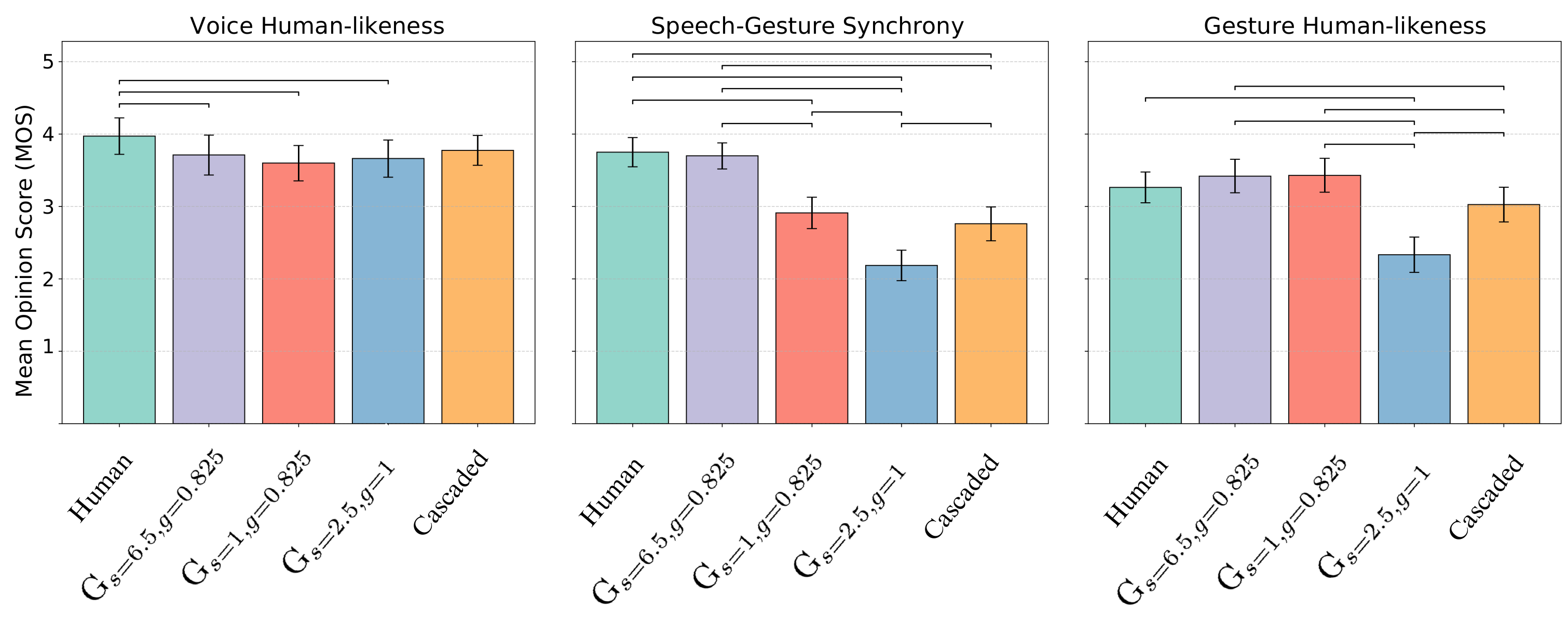}
    \caption{User study results (MOS, 95\% CI). Brackets: significant differences ($p<0.05$). \textit{Left}: voice human-likeness. \textit{Middle}: speech--gesture synchrony. \textit{Right}: gesture human-likeness.}
    \label{fig:user_study_results}
\end{figure}
Reducing gesture-history guidance significantly improves perceived synchronization and gesture human-likeness over $\mathrm{G}_{s=2.5,g=1}$, consistent with reduced motion collapse. Increasing speech guidance further improves synchronization: $\mathrm{G}_{s=6.5,g=0.825}$ obtains a score comparable to human recordings ($3.70$ versus $3.75$). Both $\mathrm{G}_{s=6.5,g=0.825}$ and $\mathrm{G}_{s=1,g=0.825}$ significantly outperform \textit{Cascaded} for gesture human-likeness. In the voice-only test, Human scores highest ($3.97$); \textit{Cascaded} obtains $3.73$ and the best-guided Gaspard $3.69$, with significant differences between \textit{Human} and some Gaspard variants. All Gaspard configurations remain comparable with the cascaded system for voice human-likeness.

\section{Discussion}
\label{sec:discussion}

The results show that the proposed method is a strong approach for expressive behavior synthesis. Gaspard outperforms unimodal gesture-generation baselines in objective and perceptual evaluations while generating \textit{both} speech and gestures. Although Gaspard generates intelligible speech, it underperforms against PocketTTS. We hypothesize that the larger number of parameters for PocketTTS (350M) compared to Gaspard (246M), and the use of Mimi, a strong audio codec, can explain these differences. Since Gaspard generates slightly lower quality speech than its cascaded counterpart, it is also possible that generating gestures jointly with speech may deteriorate speech quality. In contrast, unified generation produces highly synchronous and human-like speech and gesture behaviors, as demonstrated by the perceptual results.
Objective and perceptive results also indicate that decoupled CFG rebalances the impact of each modality during next-step prediction. Reducing $\gamma_g^G$ weakens the contribution of gesture history, mitigating dynamics collapse, while increasing $\gamma_s^G$ strengthens speech conditioning. 

\section{Conclusion and Limitations}
We presented Gaspard, a speech and gesture generation model, and we a introduced modality-decoupled CFG strategy that separately controls speech and gesture conditioning. We show that unified generation achieves leading perceptual performance, and decoupled guidance makes continuous autoregressive gesture generation expressive and mitigates dynamics collapse while improving speech-driven gesture dynamics. Compared with its cascaded counterpart, Gaspard achieves stronger gesture dynamics, perceptual gesture quality, synchronization, and lower inference time, while the cascaded system retains advantages on several unimodal objective metrics.
Limitations include the computational cost of multiple guidance passes, and evaluations of long-term generation and out-of-domain generalization remain important future work.

\section{LLM Usage Statement}

Generative AI tools were used to assist with coding, language polishing, and the summarization of research notes. During research development, we also used LLMs to generate meaningful and diverse text prompts for illustrative segments generated with Gaspard. Specifically, we used ChatGPT 5.6, 6 and Claude Sonnet 5.5. All LLM-generated outputs, including code, were manually reviewed. We did not use autonomous or agentic coding tools, such as Codex or Claude Code, during this research project.

\clearpage
\setcounter{page}{1}
\maketitlesupplementary

\section{Reproducibility statement}
In order to favor reproducibility, all code, timestamps for the evaluated samples, and trained model weights will be made available upon acceptance. 

\section{Ethical considerations}
This work primarily involved computational experiments and anonymous, non-invasive perceptual studies with consenting adult participants, for which formal ethical approval was not required under the applicable institutional guidelines. User studies were performed on the Prolific platform. Participants were all older than 18 y.o and fairly compensated. All datasets were used in accordance with their respective licenses, and the user studies were conducted with informed consent and fair participant compensation. Cloning experiments were restricted to speakers who had provided consent, and public demonstrations include safeguards intended to mitigate potential misuse.

The models together with their trained parameters will be made publicly available. While this facilitates reproducibility and enables further research, it also makes these models accessible for unintended or potentially harmful applications. In particular, because the proposed models aim to synthesize realistic human speech and behavior, they could be misused to generate convincing deepfakes, impersonate individuals, or create synthetic content without the consent of the people being represented.

\section{Dataset composition and preprocessing}
\label{app:dataset_preprocessing}

\subsection{Dataset composition}

Seamless Interaction contains $4,065$~h of paired speech--gesture data from $4,284$ participants, with associated transcriptions. Participants are recorded during dyadic interactions. The dataset is divided into two main parts: \textit{naturalistic} ($2,745$~h) and \textit{improvised} ($1,319$~h). In the naturalistic part, one participant is given a script to anchor the topic of the discussion, while the other is asked to behave naturally. This split contains spontaneous interactions but also many low-dynamics segments. In the improvised part, each participant is given a role-play prompt, leading to more expressive interactions, although less natural. The naturalistic and improvised subsets exhibit different levels of expressive motion.

Both parts are divided into training, development, and test splits. The corresponding statistics are reported in Table~\ref{tab:seamless_split}.

\begin{table}[t]
\centering
\footnotesize
\caption{Statistics of the Seamless Interaction dataset by part and split.}
\label{tab:seamless_split}
\setlength{\tabcolsep}{6pt}
\renewcommand{\arraystretch}{1.1}
\begin{tabular}{llccc}
\toprule
\textbf{Part} & \textbf{Split} & \textbf{Hours} & \textbf{Interactions} & \textbf{Participants} \\
\midrule
\multirow{3}{*}{Naturalistic}
 & Train & 2625.34 & 45,126 & 3,243 \\
 & Dev   & 35.22   & 673    & 83 \\
 & Test  & 44.13   & 786    & 109 \\
\midrule
\multirow{3}{*}{Improvised}
 & Train & 1268.01 & 16,623 & 897 \\
 & Dev   & 20.98   & 323    & 41 \\
 & Test  & 15.31   & 220    & 35 \\
\bottomrule
\end{tabular}
\end{table}

\subsection{Speech, gesture, and text preprocessing}

Speech is provided as a waveform at $48$~kHz. We resample it to $24$~kHz for data compression and to match the sample rate required by Z-Codec. Gestures are represented as full-body SMPL-H pose sequences at $30$~fps. Motion in Seamless Interaction is obtained through pseudo-mocap estimation from three simultaneous cameras. As a result, some sequences contain pose-estimation artifacts caused by occlusions, limb contacts, or interactions with surrounding objects. These artifacts can create unrealistic dynamics and discontinuities. Because all interactions were recorded in the same setting, with participants standing on a recording pod, we remove leg motion and root translation and model upper-body gestures only. Motion sequences are then resampled to $25$~fps to match the input rate of the gesture encoder.

The SMPL-H body model represents body poses as a skeleton of relative joint rotations. We convert these rotations to Rot6D~\cite{zhou2019onthecontinuity}, a representation known to be better suited to learning-based modeling. Gesture segments are therefore represented as sequences $G \in \mathbb{R}^{J \times 6 \times N}$, where $J = 43$ is the number of body joints considered and $N$ is the sequence length. Following~\cite{seamless_interaction}, we apply a Savitzky--Golay filter using a $9$-frame window and a polynomial order of $3$. This filtering reduces motion jerk without over-smoothing rapid gestures. For text, we use the provided transcriptions, obtained with WhisperX~\cite{bain2023whisperx}, together with the voice activity annotations. We do not apply the filter to evaluation segments.

\subsection{Dataset segmentation}

Segments in Seamless Interaction have variable durations, typically ranging from $2$ to $10$~min. These durations are substantially longer than the segments typically used to train TTS models, which are usually around $30$ to $60$~s. To stabilize training and improve efficiency, in particular by reducing padding and enabling length-bucketed data loading, we resegment each speech--gesture recording using the voice activity detection annotations. 

Each original recording corresponds to a full dyadic interaction, so a given participant can alternate between speaking and listening. While recent work~\cite{mughal2026miburi} shows that a model can explicitly represent this interactive state, we focus on active-speaking segments only.

\section{Model details: architecture, training and inference hyperparameters}
\label{app:training_hyperparameters}

\subsection{Z-Codec}

Z-Codec is trained on the full Seamless Interaction train set following~\cite{zcodec}. 
The complete speech codec architecture follows \cite{zcodec}, except for a latent dimension of $d_s =32$.

\subsection{Gesture $\sigma$-VAE}

\paragraph{Details about the bottleneck}
The variance structure of the latent space is critical for autoregressive next-step diffusion, since sampling uncertainty accumulates over generation steps \cite{sun2026multimodal}. However, standard VAEs are prone to variance collapse, an optimization phenomenon where the predicted variance of some channels collapses to 0, making the VAE practically deterministic in some dimensions. Variance collapse reduces robustness to exposure bias, which further harms autoregressive modeling~\cite{sun2026multimodal} and can cause variance drift or collapse that is amplified by CFG~\cite{ke2026hyperspherical}. We therefore use a variant of the vanilla VAE, termed $\sigma$-VAE, following~\cite{sun2026multimodal}. In $\sigma$-VAEs the data-dependent predicted variance of VAEs, $\sigma(x)$, is replaced by the absolute value of a variance sampled from a known normal distribution $\sigma \sim \mathcal{N}(0, C_\sigma)$. The absolute value of the sampled variance $|\sigma|$ is then fixed across channels, preventing variance collapse. The training objective of the $\sigma$-VAE becomes:
\begin{equation}
\label{eq:vae_loss}
    \mathcal{L}_{\sigma \mathrm{VAE}} = ||x - \hat{x}||_2^2 + \beta ||\mu||_2^2
\end{equation}

During training, we use the same reconstruction objective as in \cite{guichoux2026gelina}.

\paragraph{Training and architecture details}
The gesture $\sigma$-VAE is trained on the naturalistic split of Seamless Interaction for $100$k steps. We use $8$~s motion chunks, a batch size of $128$, and AdamW optimization with a learning rate decaying from $2 \times 10^{-4}$ to $4 \times 10^{-5}$. The KL-divergence weight is set to $1 \times 10^{-4}$ and $C_\sigma$ is set to $0.3$. The reconstruction objective combines geodesic, velocity, and position losses, with weights $\lambda_{\mathrm{geo}} = 1.0$, $\lambda_{\mathrm{vel}} = 1.0$, and $\lambda_{\mathrm{pos}} = 1.0$.

\subsection{Gaspard}

\begin{table}[t]
\centering
\footnotesize
\caption{Frame rates and dimensionality reduction across the speech and gesture pipelines used by Gaspard.}
\label{tab:framerate_dims}
\setlength{\tabcolsep}{5pt}
\renewcommand{\arraystretch}{1.1}
\resizebox{\linewidth}{!}{%
\begin{tabular}{lllcc}
\toprule
\textbf{Modality} & \textbf{Stage} & \textbf{Module} & \textbf{Frame rate} & \textbf{Latent dim.} \\
\midrule
\multirow{3}{*}{Speech}
 & Input waveform & N/A      & 24\,kHz  & N/A \\
 & Stage 1        & WavVAE   & 100\,Hz  & 24 \\
 & Stage 2        & PatchVAE & 12.5\,Hz & 32 \\
\midrule
\multirow{2}{*}{Gesture}
 & Input sequence & Rot6D        & 25\,fps  & N/A \\
 & Encoding       & $\sigma$-VAE & 12.5\,Hz & 64 \\
\midrule
\multirow{3}{*}{Both}
 & Projection & $\phi_{\mathrm{in}}^s$ / $\phi_{\mathrm{in}}^g$ & 12.5\,Hz & $32/64 \rightarrow 1024$ \\
 & Backbone & Gaspard decoder & 12.5\,Hz & 1024 \\
 & Flow-matching heads & $h_\theta^s$ / $h_\theta^g$ & 12.5\,Hz & $1024 \rightarrow 32/64$ \\
\bottomrule
\end{tabular}}

\end{table}

Gaspard is trained for $200$k steps on $2$ H100 GPUs, corresponding to approximately $40$~h of training. We use AdamW with a learning rate decaying from $1 \times 10^{-4}$ to $1 \times 10^{-5}$ and batches of $20$k embeddings. The speech flow-matching loss weight is set to $\lambda_s = 1.0$. The gesture loss weight is linearly increased from $1 \times 10^{-4}$ to $\lambda_g = 1.3$ over the first $70$k steps after warmup. The gesture loss is ramped because text--speech alignment emerges first: without this ramp, increasing gesture supervision too early leads to speech underfitting. The stop-prediction loss is weighted by $0.05$.

During Gaspard training, text, speech, and gesture conditions are randomly dropped with probabilities $0.01$, $0.05$, and $0.05$, respectively. The context vector output is independently and randomly dropped with probability $0.1$, in which case the conditioning of the flow-matching head is a zero-vector.
We also apply latent noise perturbation to speech and gesture inputs. The amount of noise is sampled from beta distributions with parameters $(10,1)$ for speech and $(15,1)$ for gestures.

\subsubsection{Detailed discussion about CFG}
\label{app:cfg_formulation}

One can recover the additive formulation of multimodal CFG with reparameterization, as follows:
\begin{equation}
\begin{aligned}
v_{\mathrm{cfg}} &\leftarrow v,\\
v_{\mathrm{cfg}} &\leftarrow v^{\neg G}+\gamma_g^m\left(v_{\mathrm{cfg}}-v^{\neg G}\right),\\
v_{\mathrm{cfg}} &\leftarrow v^{\neg S}+\gamma_s^m\left(v_{\mathrm{cfg}}-v^{\neg S}\right).
\end{aligned}
\end{equation}

For fixed guidance scales, this is equivalent to
\begin{equation}
v_{\mathrm{cfg}}
=
v
+
\gamma_s^m(\gamma_g^m-1)(v-v^{\neg G})
+
(\gamma_s^m-1)(v-v^{\neg S}).
\end{equation}

For gesture generation, with $\gamma_g^G=0.825$ and $\gamma_s^G=6.5$,
\begin{equation}
v_{\mathrm{cfg}}
=
v
-
1.1375(v-v^{\neg G})
+
5.5(v-v^{\neg S}).
\end{equation}

We present the intuition behind the different CFG coefficients in Table~\ref{tab:cfg_coefficients}.
\begin{table}[!htb]
    \centering
    \small
    \caption{Interpretation of the modality-specific CFG coefficients.}
    \label{tab:cfg_coefficients}
    \setlength{\tabcolsep}{5pt}
    \renewcommand{\arraystretch}{1.2}
    \begin{tabularx}{\linewidth}{@{}l c X@{}}
        \toprule
        \textbf{Head} & \textbf{Coefficient} & \textbf{Interpretation} \\
        \midrule
        \multirow{2}{*}{Speech head $h_\theta^S$}
        & $\gamma_s^S$ & Controls the strength of \textit{speech} conditioning when predicting the next-step \textit{speech} embedding. \\
        & $\gamma_g^S$ & Controls the strength of \textit{gesture} conditioning when predicting the next-step \textit{speech} embedding. \\
        \midrule
        \multirow{2}{*}{Gesture head $h_\theta^G$}
        & $\gamma_s^G$ & Controls the strength of \textit{speech} conditioning when predicting the next-step \textit{gesture} embedding. \\
        & $\gamma_g^G$ & Controls the strength of \textit{gesture} conditioning when predicting the next-step \textit{gesture} embedding. \\
        \bottomrule
    \end{tabularx}
\end{table}

\subsubsection{Inference}
\begin{algorithm*}[t]
\caption{Autoregressive inference with Gaspard. Each autoregressive step requires three backbone inferences to obtain the full, gesture-masked, and speech-masked contexts. Speech and gesture next-step embeddings are generated through modality-specific flow-matching ODE solves with decoupled classifier-free guidance. Generated gesture embeddings are reused as autoregressive context.}
\label{alg:gaspard_inference}

\KwIn{Input text $x$, number of autoregressive steps $T$, number of flow-matching sampling steps $S_{\mathrm{ODE}}$, guidance scales $\gamma_g^G$, $\gamma_s^G$, $\gamma_g^S$, $\gamma_s^S$}
\KwOut{Generated speech and gesture sequences}

Initialize $z^s_0$ and $z^g_0$\;

\For{$t = 0, \ldots, T-1$}{

    \tcp{Compute autoregressive contexts}

    $c_{t+1}
    =
    B_\theta
    \left(
    \mathrm{Fuse}(z^s_{\leq t}, z^g_{\leq t}), x
    \right)$
    \tcp*[r]{fully conditioned}

    $c_{t+1}^G
    =
    B_\theta
    \left(
    \mathrm{Fuse}(z^s_{\leq t}, \varnothing), x
    \right)$
    \tcp*[r]{gesture masked}

    $c_{t+1}^S
    =
    B_\theta
    \left(
    \mathrm{Fuse}(\varnothing, z^g_{\leq t}), x
    \right)$
    \tcp*[r]{speech masked}

    \tcp{Generate the next continuous speech embedding with decoupled CFG}
    $z^s_{t+1}
    =
    \mathrm{ODESolve}
    \left(
    h_\theta^s,
    c_{t+1},
    c_{t+1}^{\neg G},
    c_{t+1}^{\neg S},
    \gamma_g^S,
    \gamma_s^S,
    S_{\mathrm{ODE}}
    \right)$\;

    \tcp{Generate the next continuous gesture embedding with decoupled CFG}
    $z^g_{t+1}
    =
    \mathrm{ODESolve}
    \left(
    h_\theta^g,
    c_{t+1},
    c_{t+1}^{\neg G},
    c_{t+1}^{\neg S},
    \gamma_g^G,
    \gamma_s^G,
    S_{\mathrm{ODE}}
    \right)$\;

}

\tcp{Decode the generated latent sequences into their output modalities}
Decode $z^s_{0:T}$ into waveform\;

Decode $z^g_{0:T}$ into motion\;

\KwRet{generated speech and gesture sequences}\;

\end{algorithm*}

At inference time, we use $15$ flow-matching sampling steps for both the speech and gesture heads. The sampling temperature is set to $1.0$ for the speech head and $0.975$ for the gesture head, and trajectories are integrated using the Euler ODE solver. For classifier-free guidance, we use $\gamma_g^G = 0.825$, $\gamma_s^G = 6.5$, $\gamma_g^S = 1.3$, and $\gamma_s^S = 1.3$, unless specified otherwise. These values were empirically set after perceptual inspection of generated segments on the \textit{dev} set, and fixed for subsequent evaluation.

\subsection{Cascaded baseline}
\label{app:baselines}

\begin{figure}[!htb]
    \centering
    \includegraphics[width=0.6\linewidth]{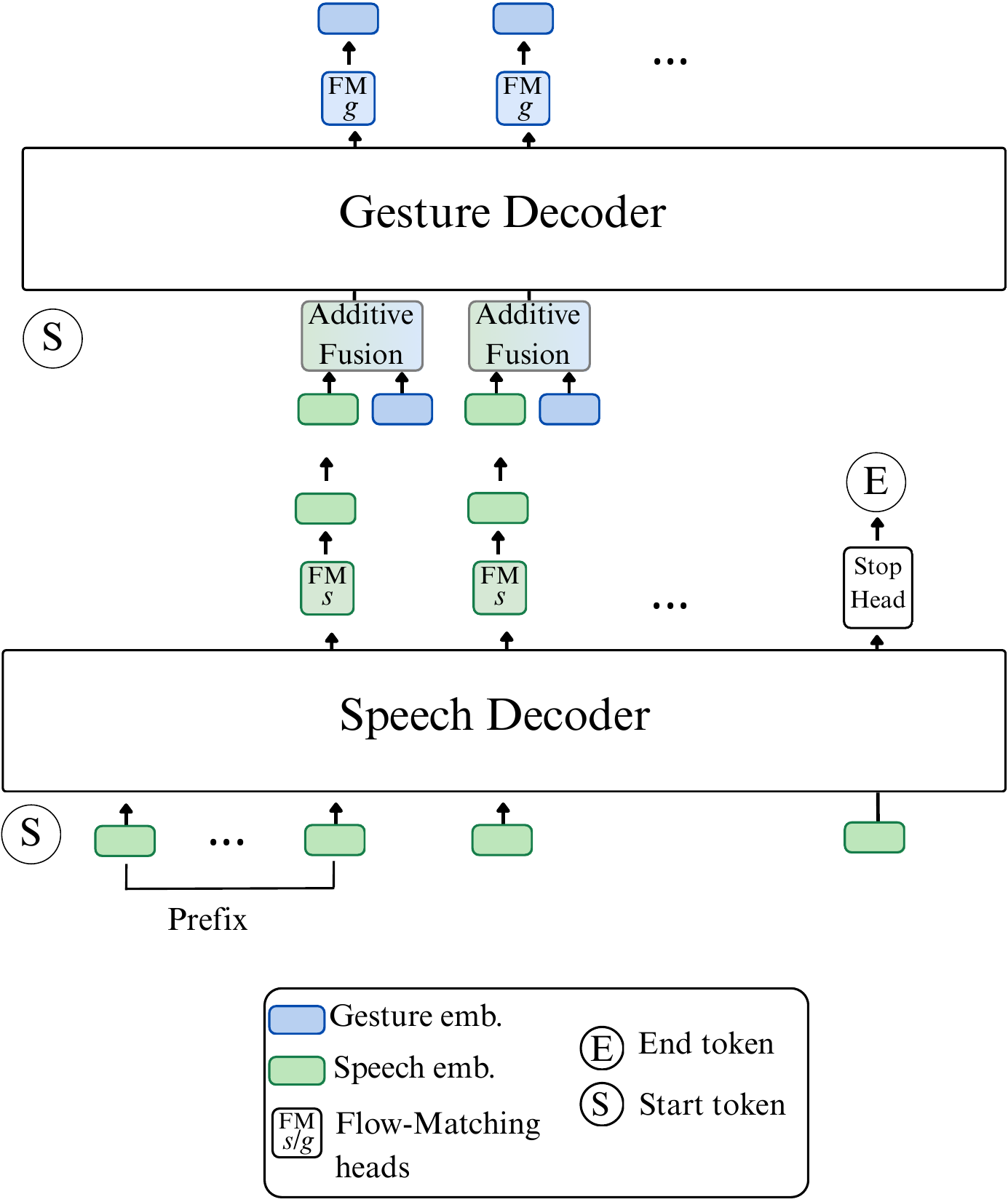}
    \caption{Simplified overview of the cascaded TTS and speech-driven gesture generation baseline. The TTS module first generates speech conditioned on text, after which the resulting speech embeddings are provided as input to the gesture generation module. Speech synthesis is therefore performed independently of gesture generation, as in most conventional cascaded systems.}
    \label{fig:cascaded}
\end{figure}

Figure~\ref{fig:cascaded} presents a simplified overview of the \textit{Cascaded} system.
The TTS component of \textit{Cascaded} corresponds exactly to Gaspard without the gesture input projections and flow-matching head, it is therefore a 222M~parameter model. The gesture generation component is a 246M~parameter generation model, mimicking Gaspard.

For inference, we used $\gamma_s^S = 1.3$, $\gamma^G_g=0.825$ and $\gamma^G_s=6.5$, as in Gaspard.

\section{Evaluation metric implementation and validation}
\label{app:metric_details}

\subsection{Kinetic FGD}
 Given a sequence of extracted features
\begin{equation}
    F = \left(f_1, \ldots, f_N\right),
\end{equation}
we first compute its latent velocity sequence as
\begin{equation}
    \Delta f_t = f_{t+1} - f_t,
    \qquad t \in \{1, \ldots, N-1\}.
\end{equation}

Each gesture sequence is then represented by a kinetic descriptor obtained by concatenating the absolute temporal mean and standard deviation of its latent velocities:
\begin{equation}
    v(F)
    =
    \left[
        \operatorname{mean}_{t}\left(|\Delta f_t|\right),
        \operatorname{std}_{t}\left(\Delta f_t\right)
    \right].
\end{equation}

Let $\mathcal{V}_{\mathrm{GT}}$ and $\mathcal{V}_{\mathrm{pred}}$ denote the sets of kinetic descriptors extracted from the ground-truth and generated sequences, respectively. Kinetic FGD is defined as
\begin{equation}
    \mathrm{FGD}_k
    =
    D_{\mathrm{Fr\acute{e}chet}}
    \left(
        \mathcal{V}_{\mathrm{GT}},
        \mathcal{V}_{\mathrm{pred}}
    \right).
\end{equation}

\subsection{FGD feature extractor}

Unlike BEAT2, Seamless Interaction does not provide a pretrained feature extractor suitable for computing FGD. We therefore train a non-causal VAE. The feature extractor encodes gesture sequences sampled at $30$~fps into continuous latent embeddings at $7.5$~Hz, with a latent dimensionality of $40$. The architecture is not primarily designed to maximize frame-level reconstruction accuracy. Instead, its temporal compression and relatively compact latent space encourage the aggregation of motion information into semantically meaningful representations, which is more appropriate for distribution-level evaluation with FGD. The feature extractor is trained on the \textit{improvised} training split, and the evaluated segments are not seen during training.

Conventional FGD compares the distribution of extracted latent features but does not explicitly characterize how these features evolve over time. Two sets of sequences may therefore exhibit similar feature distributions while differing substantially in motion amplitude, smoothness, or temporal variability. This limitation is particularly relevant for autoregressive models, which may produce realistic individual poses while suffering from over-smoothed, static, or locally unstable motion dynamics.

\subsection{Validation of kinetic FGD}

To validate the sensitivity of $\mathrm{FGD}_k$ to the motion dynamics, we computed FGD and $\mathrm{FGD}_k$ on the clean test set after applying controlled perturbations to the original gesture sequences:
\begin{itemize}
    \item We modified the velocity of gesture sequences through temporal interpolation or downsampling.
    \item We smoothed the original gesture sequences using a Savitzky-Golay filter, and varied the filter window size to progressively increase the smoothness.
    \item We altered the original gesture sequences by adding a jerk component, defined as the difference between the original and smoothed motion and scaled by a factor controlling the perturbation magnitude.
\end{itemize}

\begin{figure}[!htb]
    \centering
    \includegraphics[width=0.8\linewidth]{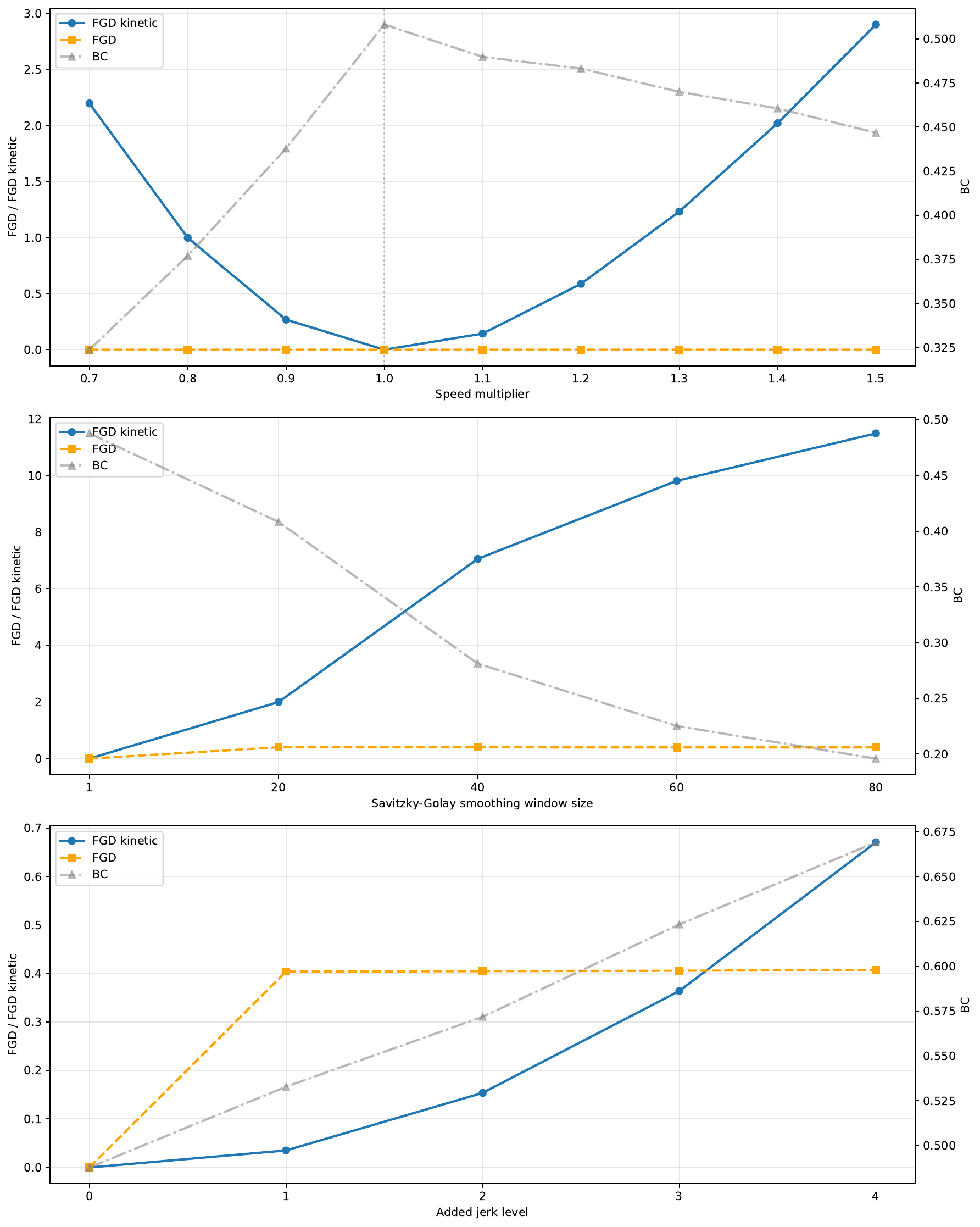}
    \caption{We validate the $\mathrm{FGD}_k$ metric by varying gesturing velocity (\textit{Top}), smoothness (\textit{Middle}) and jerk (\textit{Bottom}). We also include the BC score corresponding to all three perturbations. FGD shows little sensitivity to the perturbations, whereas $\mathrm{FGD}_k$ captures velocity, smoothness and jerk variations.}
    \label{fig:fgd_k}
\end{figure}

As shown in Figure~\ref{fig:fgd_k}, changes in gesture speed, smoothness and jerk have little effect on FGD, whereas ${\mathrm{FGD}_k}$ is much more sensitive to these temporal variations. Notably, FGD remains very low despite modified gesture velocity.

\subsection{Beat Consistency calibration}

Because the motion dynamics and pose distributions of Seamless Interaction differ substantially from those of BEAT2, the hyperparameters of the Beat Consistency metric require dataset-specific calibration. We adjust the velocity--acceleration threshold used for gesture-stroke detection and the parameter $\sigma$, which controls the temporal tolerance of the alignment function. These parameters are empirically set to $0.25$ and $0.3$, respectively. We select this configuration by maximizing the difference in BC score between correctly aligned speech--gesture pairs and artificially misaligned pairs, thereby ensuring that the metric discriminates between synchronized and unsynchronized sequences on Seamless Interaction.

A limitation of BC is that temporally unstable or jerky motion may also produce a high score, as rapid motion variations increase the number of detected gesture beats and therefore the likelihood of alignment with speech beats. BC should consequently not be interpreted independently, but jointly with FGD and $\mathrm{FGD}_k$, which respectively evaluate the realism of the gesture representations and their latent dynamics. We illustrate this effect in Figure~\ref{fig:fgd_k}.

\subsection{WER}
For WER computation, we used WhisperX \citep{bain2023whisperx} to obtain speech transcriptions and applied peak normalization to the input audio. Both the ground-truth recordings and the generated samples were transcribed using this ASR. Since the speech segments from the Seamless Interaction dataset originate from spontaneous dyadic interactions, it is important to note that the reference transcriptions may contain inconsistencies. In particular, some ground-truth recordings contain speech leakage (i.e. the interlocutor's speech) or background noise from the interlocutor, which may also be transcribed by the ASR system. This difference from the clean, studio-recorded data commonly used for TTS evaluation partly explains why the reported WER scores remain relatively high compared with state-of-the-art TTS systems.

\section{Evaluation protocol details}

\subsection{Construction of the Seamless-416 evaluation subset}
\label{app:clean_subset}
\label{app:prefix_selection}

Since the original Seamless Interaction test split is relatively large (approximately $60$~h), we construct a smaller subset for evaluation. The candidate subset contains recordings ranging from $15$ to $35$~s and initially covers $103$ speakers from the raw test set of Seamless Interaction. For each speaker, we additionally select a random shorter segment with a maximum duration of $10$~s, which is used as a speech and gesture prefix for Gaspard generation. These prefixes do not overlap with the evaluated segments.
A substantial number of candidate segments present limited motion or insufficient speech quality. To ensure a fair and meaningful evaluation, we initially sample $487$ segments from the test split, corresponding to approximately four to six segments per speaker. We then exclude unsuitable segments and prefixes according to the following criteria:
\begin{itemize}
    \item We retain only segments with a static gesture ratio below $50\%$, ensuring that evaluated sequences are non-static and contain meaningful gestures.
    \item We discard segments containing unintelligible speech caused by recording issues.
    \item We discard segments for which the provided transcription does not match the recorded speech.
\end{itemize}

This screening is performed exclusively on ground-truth human recordings and is not based on the performance of the evaluated models. The resulting clean test subset contains $416$ segments from $93$ speakers, corresponding to a total duration of approximately $2$~h~$35$~min. The complete list of selected files and their corresponding timestamps will be made publicly available to support reproducibility.

\subsection{Seamless-416 user study}
We constructed the user study evaluation set by randomly sampling $13$ speakers from the \textit{Seamless-416} subset. For each of the speakers, we selected up to four segments of $15$--$35$ seconds, discarding segments containing pose glitches or unintelligible speech; this screening is performed solely based on the recordings from the dataset. The resulting evaluation set consists of $24$ segments from $11$ speakers, including $4$ males and $7$ females.

All evaluated systems used their generated speech and gestures. Differences in perceived synchrony may therefore originate from variations in speech and variations in gestures. Isolating either, by fixing one modality input, is possible but out of the scope of the present research and is left for future investigations.

\subsection{Genea-95 evaluation subset}
\label{app:genea_90}
The Genea Challenge 2026 performed a benchmark evaluation on a manually curated subset of the test set of Seamless Interaction. The evaluation set provided by the organizers contains $95$ segments in the \textit{public} subset and $83$ in the \textit{private} subset. 
Because the human recordings are not available for the \textit{private} subset, we only consider the \textit{public} one.
For each speaker, we manually selected a short segment with a maximum duration of $10$~s from the test set of the Seamless Interaction dataset. We discarded prefixes containing glitches or unintelligible speech. The prefixes do not overlap with the evaluated segments.
There is no overlap between the \textit{Genea-95} and the \textit{Seamless-416} evaluation sets.

\paragraph{Fairness of prefix conditioning}
Because Gaspard is conditioned by a short speech-gesture prefix, the synthesis may contain more speaker-specific information than the baselines. We do acknowledge that it might affect the fairness of the evaluation. Though, this conditioning can be seen as a sort of speaker-embedding, a common strategy in gesture generation systems. Prefix conditioning is also necessary to make the evaluation more controlled, allowing to clone a target speaker voice allowing fairer comparisons against speech-driven baselines.

\subsubsection{Genea-95 user study}
For the perceptual evaluation, we randomly sampled $10$ different speakers ($7$ females and $3$ males) contained in the \textit{Genea-95} subset. For each speaker, we then randomly sampled segments until reaching a total of $30$ segments.

\subsection{Genea baselines}
We performed our evaluation against $4$ gesture generation baselines: \textit{UNICAMP} \cite{sanchez2026unified}, \textit{UM-FERI} \cite{crnek2026umferi}, \textit{GestFlow} \cite{guillen2026gestflow} and \textit{DyaSync-Sem} \cite{fang2026dyasync}, which correspond to the submissions at the Genea Challenge 2026 \cite{nagy2026genea}. Although the Genea Challenge also evaluated \textit{DyaSync}, a variant of \textit{DyaSync-Sem}, we did not retain that model for cost management reasons, and only retained \textit{DyaSync-Sem} as the best performing model between the two.

We did \textit{not} run the inferences for all models, we instead used the segments submitted to the challenge, which were provided by the organizers. More specifically, we used the matched speech-gesture monadic segments included in the public set.

\section{Complementary results}

\subsection{Inference efficiency}
\label{sec:sampling_steps}

\begin{figure}[!htb]
    \centering
    \includegraphics[width=0.65\linewidth]{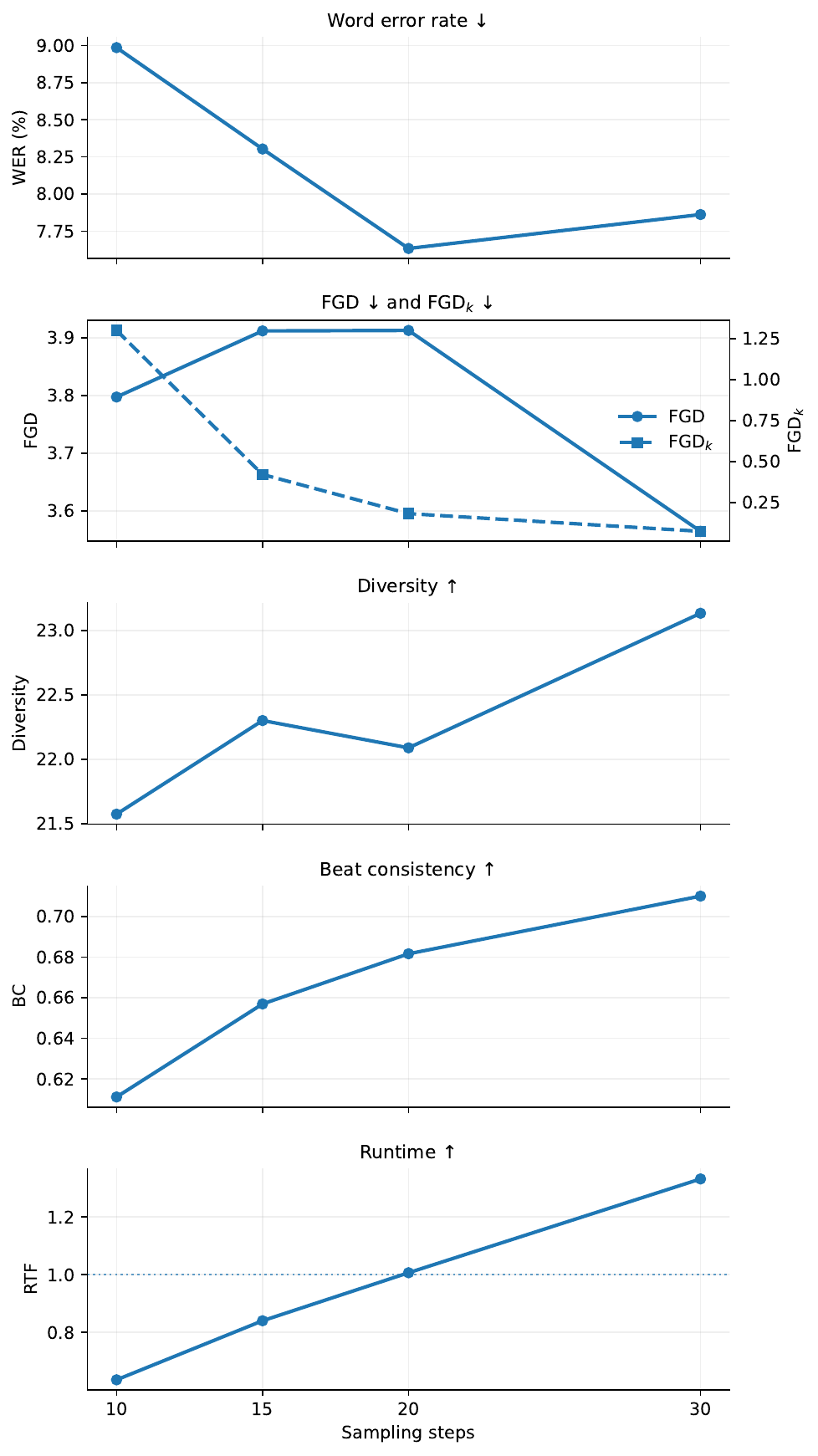}
    \caption{Quality--efficiency trade-off with respect to the number of flow-matching sampling steps. Increasing the number of sampling steps consistently improves generation performance at the cost of lower inference speed.}
    \label{fig:sampling_steps}
\end{figure}

We investigate Gaspard's inference efficiency by varying the number of flow-matching sampling steps. Figure~\ref{fig:sampling_steps} shows that increasing the number of sampling steps generally improves both speech and gesture generation quality, at the cost of increased inference time. Increasing the number of steps from $10$ to $30$ raises the RTF from $0.63$ to $1.33$. In all experiments, we therefore use $15$ sampling steps as a compromise between generation quality and real-time efficiency, corresponding to an RTF of $0.84$.

Decoupled CFG, however, requires more computation because multiple autoregressive streams are needed to access the different conditioning variants. This directly impacts inference speed, as standard CFG achieves an average RTF of $0.78$, compared with $0.84$ for our decoupled formulation. These results are described in Appendix~\ref{app:standard_cfg}.

\subsection{Comparison to standard CFG}
\label{app:standard_cfg}
We define standard CFG as follows:
$$
v_{\mathrm{cfg}}^m = v_{\varnothing}^m + \gamma^m (v^m - v_{\varnothing}^m), ~~~m\in \{S,G\}
$$
Where $v_\varnothing^m = h^m_\theta(\varnothing, z_\tau, \tau)$ corresponds to the predicted unconditional velocity field by the flow-matching head, and $v^m = h^m_\theta(c_t, z_\tau, \tau)$ is the fully conditional prediction.

To complement the comparison between standard and decoupled CFG presented in Table~\ref{tab:cfg_comparison} of the main paper, we report additional results on the selection of the guidance scales used for the standard CFG baseline. Table~\ref{tab:standard_cfg_sweep} studies the gesture guidance scale $\gamma^G$, while Table~\ref{tab:standard_cfg_speech_sweep} studies the speech guidance scale $\gamma^S$. The best overall gesture generation objective results are obtained with $\gamma^G=0.75$, although $\gamma^G = 0.85$ leads to better FGD, it produces less stable and realistic dynamics.

\begin{table}[!htb]
\centering
\footnotesize
\caption{Impact of the standard classifier-free guidance scale on gesture generation performance.}
\label{tab:standard_cfg_sweep}
\setlength{\tabcolsep}{5pt}
\renewcommand{\arraystretch}{1.10}

\begin{tabularx}{\linewidth}{c *{4}{>{\centering\arraybackslash}X}}
\toprule
$\boldsymbol{\gamma^{G}}$
& \textbf{FGD} $\downarrow$
& $\mathbf{FGD}_{k}$ $\downarrow$
& \textbf{BC} $\sim$
& \textbf{Div.} $\sim$ \\
\midrule
0.75

& 4.34 $\pm$ 1.18
& 0.81$\pm$0.43
& 0.761$\pm$0.05
& 22.02$\pm$0.83 \\

0.85
& 2.96 $\pm$ 0.26
& 3.14 $\pm$ 0.25
& 0.361 $\pm$ 0.02
& 15.93 $\pm$ 0.86 \\

1.00
& 8.63 $\pm$ 0.66
& 12.53 $\pm$ 0.18
& 0.050 $\pm$ 0.01
& 11.29 $\pm$ 0.84 \\

1.50
& 11.41 $\pm$ 0.18
& 11.72 $\pm$ 0.21
& 0.127 $\pm$ 0.02
& 15.11 $\pm$ 0.26 \\

2.50
& 10.52 $\pm$ 0.17
& 11.00 $\pm$ 0.54
& 0.156 $\pm$ 0.02
& 12.97 $\pm$ 0.58 \\
\bottomrule
\end{tabularx}

\end{table}

\begin{table}[!htb]
\centering
\footnotesize
\caption{Impact of the standard classifier-free speech guidance scale $\gamma^S$ on speech generation performance, with $\gamma^G=0.75$ fixed.}
\label{tab:standard_cfg_speech_sweep}
\setlength{\tabcolsep}{5pt}
\renewcommand{\arraystretch}{1.10}

\begin{tabularx}{\linewidth}{c *{3}{>{\centering\arraybackslash}X}}
\toprule
$\boldsymbol{\gamma^{S}}$
& \textbf{WER} $\downarrow$
& \textbf{UTMOS} $\uparrow$
& \textbf{SS} $\uparrow$ \\
\midrule
1.0
& 21.04 $\pm$ 0.98
& 1.91 $\pm$ 0.39
& 0.74 $\pm$ 0.01 \\

1.1
& 21.55 $\pm$ 1.22
& 1.91 $\pm$ 0.02
& 0.74 $\pm$ 0.01 \\

1.2
& 21.56 $\pm$ 1.21
& 1.90 $\pm$ 0.02
& 0.74 $\pm$ 0.01 \\

1.3
& 21.57 $\pm$ 1.02
& 1.91 $\pm$ 0.02
& 0.74 $\pm$ 0.01 \\
\bottomrule
\end{tabularx}

\end{table}

\subsection{Speech adaptation to gesture}
\label{app:speech_adaptation}
One key motivation for generating speech and gesture jointly is to allow the generation of each modality to be conditioned on the other. To analyze the influence of gestures on speech generation, we conducted controlled experiments in which a ground-truth gesture sequence was imposed on the gesture stream while the model generated the corresponding speech. We then modified the kinematic properties of this conditioning gesture input, in particular its speed, while keeping the input text unchanged. This setting allows us to examine how changes in gesture kinematics influence the generated speech. We did not constrain the duration of the generated speech. For each condition, generation ends once the complete input text has been synthesized. If the input gesture sequence ends before speech generation is complete, the model resumes joint autoregressive generation of both speech and gesture until the full input text is generated.

For this experiment, we selected $31$ samples from the test set, covering $8$ different speaker identities. For each sample, we generated two speech sequences using two different random seeds and the same input text. Speech generation was conditioned on temporally scaled versions of the corresponding gesture sequence, using speed multipliers in $[0.8, 0.9, 1, 1.1, 1.2]$. As a result, each speed multiplier corresponds to $62$ generated speech sequences, obtained from $31$ distinct gesture--text pairs.

\begin{figure}[!htb]
    \centering
    \begin{subfigure}[t]{0.49\linewidth}
        \centering
        \includegraphics[width=\linewidth]{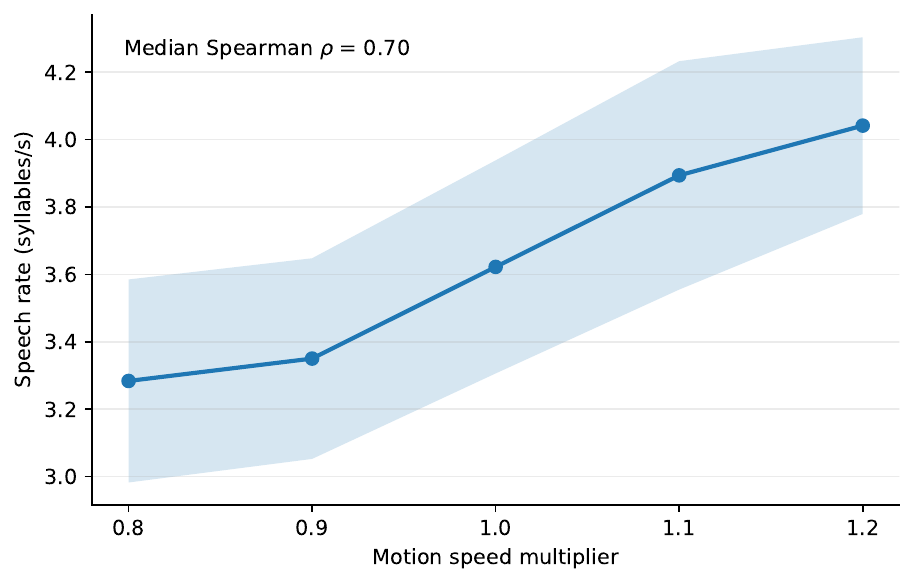}
        \caption{Impact of gesture speed on speech rate.}
        \label{fig:speech_rate_speed}
    \end{subfigure}
    \hfill
    \begin{subfigure}[t]{0.49\linewidth}
        \centering
        \includegraphics[width=\linewidth]{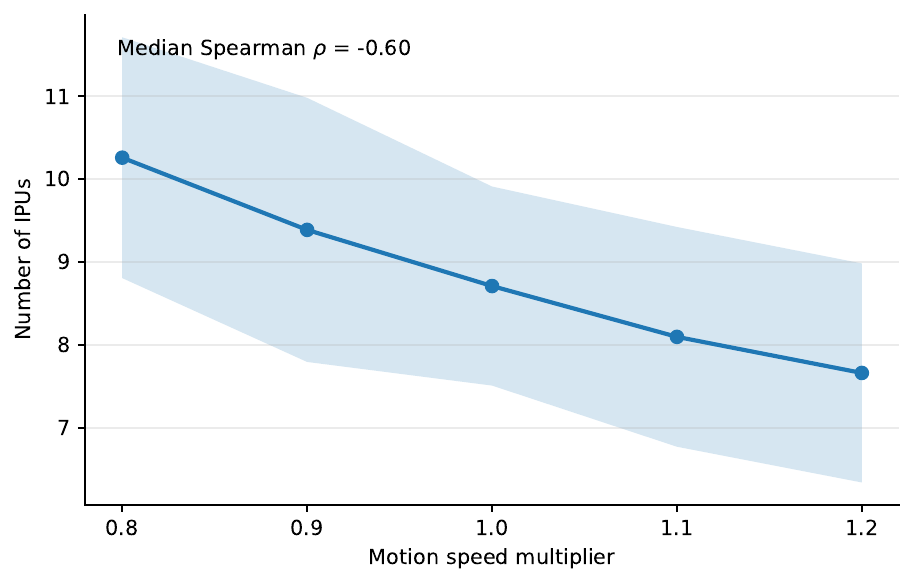}
        \caption{Impact of gesture speed on the number of interpausal units (IPUs).}
        \label{fig:speech_ipu}
    \end{subfigure}
    \caption{Impact of gesture speed on speech generation. Speech is generated while conditioning the model on fixed gesture inputs with varying speeds. Increasing gesture speed results in a higher speech rate and fewer IPUs.}
    \label{fig:speech_rate}
\end{figure}

We report the evolution of the average speech rate and inter-pausal unit (IPU) number, together with their corresponding median Spearman correlation coefficient, in Figure~\ref{fig:speech_rate}. Speech rate is computed as the number of syllables per second. An IPU corresponds to a stretch of speech boudned by pauses. An IPU is detected when a speaker remains silent for at least $0.2$~s. Progressively increasing gesturing speed increases the average speech rate from $3.27$~syllables/s to $4.03$~syllables/s while decreasing the average number of IPUs from around $10$ to below $8$. Both the variations in speech rate and IPU number exhibit a fairly strong correlation as indicated by a median Spearman correlation coefficient of $\rho = 0.7$ and $\rho = -0.6$ respectively.

These median coefficients, however, do not fully account for the inter-speaker variability. We therefore report per-sample Spearman correlation in Table~\ref{tab:speech_rate_ipu_spearman}.
\begin{table}[!htb]
    \centering
    \footnotesize
    \caption{Spearman correlations between gesture speed multiplier and generated speech characteristics for each sample.}
    \label{tab:speech_rate_ipu_spearman}
    \setlength{\tabcolsep}{7pt}
    \renewcommand{\arraystretch}{0.95}
    \begin{tabular}{cc|c|c}
        \toprule
        \textbf{Sample ID} & \textbf{Speaker}
        & \textbf{Speech rate}
        & \textbf{Number of IPUs} \\
        & & \textbf{Spearman $\rho$}
        & \textbf{Spearman $\rho$} \\
        \midrule
         1 & 0093 & $1.000$ & $-0.800$ \\
         2 & 0093 & $0.200$ & $-0.718$ \\
         3 & 0093 & $0.200$ & $-0.975$ \\
         4 & 0093 & $-0.100$ & $0.700$ \\
         5 & 0093 & $0.600$ & $-0.600$ \\
         6 & 0222 & $0.600$ & $-0.300$ \\
         7 & 0222 & $0.800$ & $-0.308$ \\
         8 & 0222 & $1.000$ & $0.462$ \\
         9 & 0222 & $-0.100$ & $-0.667$ \\
        10 & 0222 & $0.800$ & $-0.400$ \\
        11 & 0561 & $0.100$ & $-0.900$ \\
        12 & 0561 & $0.700$ & $-0.671$ \\
        13 & 0561 & $1.000$ & $-0.975$ \\
        14 & 0561 & $0.300$ & $-0.500$ \\
        15 & 0739 & $0.800$ & $-1.000$ \\
        16 & 0739 & $0.100$ & $-0.205$ \\
        17 & 0739 & $0.600$ & $-0.872$ \\
        18 & 0739 & $0.700$ & $-0.667$ \\
        19 & 0742 & $0.300$ & $-0.200$ \\
        20 & 0831 & $0.900$ & $-0.900$ \\
        21 & 0831 & $0.900$ & $-0.600$ \\
        22 & 0831 & $0.700$ & $-0.800$ \\
        23 & 0831 & $0.500$ & $-0.205$ \\
        24 & 1144 & $0.800$ & $-0.462$ \\
        25 & 1144 & $0.800$ & $-0.308$ \\
        26 & 1144 & $0.500$ & $-0.211$ \\
        27 & 1144 & $0.900$ & $-0.564$ \\
        28 & 5124 & $0.700$ & $-0.300$ \\
        29 & 5124 & $0.600$ & $-0.400$ \\
        30 & 5124 & $0.600$ & $-0.700$ \\
        31 & 5124 & $0.900$ & $-0.707$ \\
        \bottomrule
    \end{tabular}
\end{table}

\subsection{Qualitative results}
\begin{figure}[!htb]
    \centering
    \includegraphics[width=1.0\linewidth]{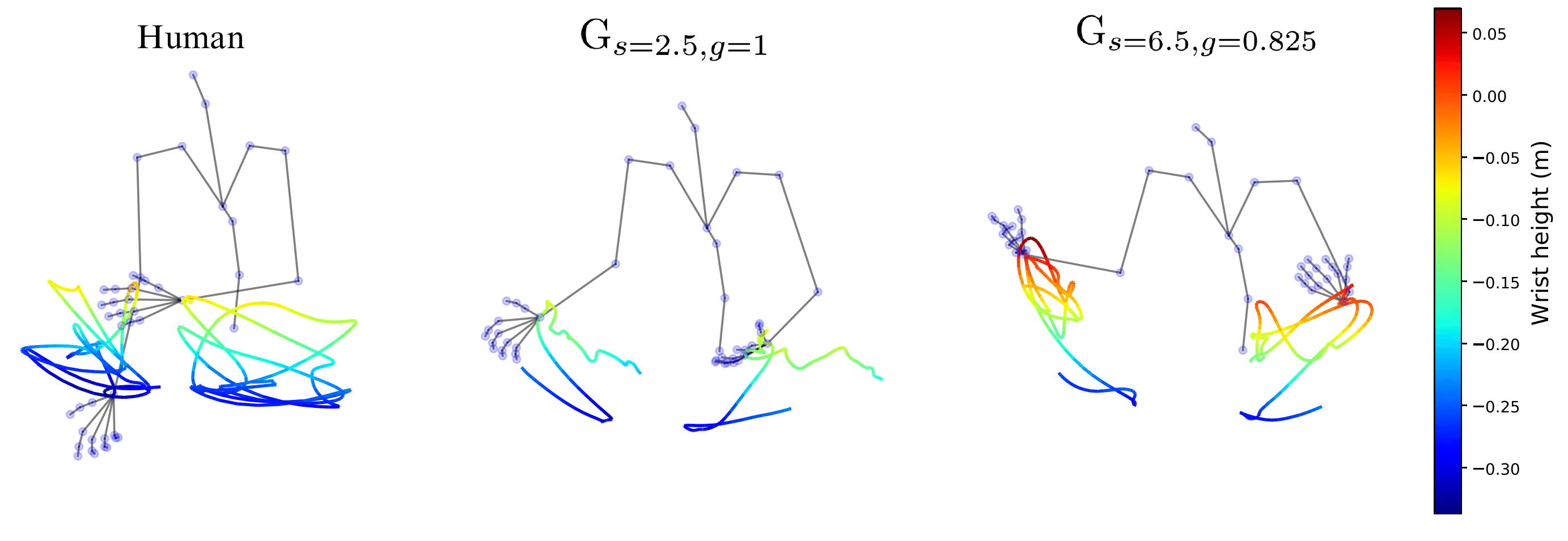}
    \caption{Example wrist trajectories for three stimuli used in the \textit{Seamless-416} user study with different CFG configurations. \textit{Left:} Human recording. \textit{Middle:} Gesture sequence generated by setting $\gamma^G_g=1$, $\gamma^G_s=2.5$. \textit{Right:} Gesture sequence generated by setting $\gamma^G_g=0.825$, $\gamma^G_s=6.5$. Reducing $\gamma^G_g$ while increasing $\gamma^G_s$ increases the amplitude and variability of the generated gestures.}
    \label{fig:wrists}
\end{figure}

We provide a qualitative illustration of the effect of CFG coefficient variations on the dynamics of the generated gestures in Figure~\ref{fig:wrists}. All three trajectories correspond to stimuli evaluated during the user study.
Reducing the guidance with respect to the gesture history to $\gamma_g^G = 0.825$, while increasing the guidance with respect to speech conditioning to $\gamma_s^G = 6.5$, results in larger and more variable wrist trajectories.

{
    \small
    \bibliographystyle{ieeenat_fullname}
    \bibliography{main}
}


\end{document}